%% file: Preprint.tex
\documentclass[letterpaper]{article} 
\usepackage{aaai2027}  
\usepackage[hyphens]{url}  
\usepackage{graphicx} 
\usepackage{natbib}  
\usepackage{caption} 
\usepackage{algorithm}
\usepackage{algorithmic}
\usepackage{bbding}
\usepackage{algorithm}
\usepackage{algorithmic}
\usepackage{amsfonts}
\usepackage{amssymb}
\usepackage{amsmath}
\usepackage{booktabs}

\usepackage[most]{tcolorbox}

\newtcolorbox{questionbox}[1][]{
  enhanced,
  colback=red!3,
  colframe=red!40!black,
  boxrule=0.5pt,
  arc=2mm,
  left=1.5mm,
  right=1.5mm,
  top=1mm,
  bottom=1mm,
  fonttitle=\bfseries,
  title=#1
}

\newtcolorbox{insightbox}[1][]{
  enhanced,
  colback=blue!3,
  colframe=blue!40!black,
  boxrule=0.5pt,
  arc=2mm,
  left=1.5mm,
  right=1.5mm,
  top=1mm,
  bottom=1mm,
  fonttitle=\bfseries,
  title=#1
}

\definecolor{cBaseline}{HTML}{cfcfcf}
\definecolor{cRepa}{HTML}{a1c9f4}
\definecolor{cVF}{HTML}{ffb482}
\definecolor{cMcos}{HTML}{8de5a1}
\definecolor{cMdms}{HTML}{ff9f9b}
\definecolor{cEq}{HTML}{d0bbff}
\definecolor{cLcr}{HTML}{debb9b}
\definecolor{cLmr}{HTML}{fab0e4}
\definecolor{cMae}{HTML}{fffea3}

\newcommand{\clusterDot}[1]{%
  \tikz[baseline=-0.7ex]\draw[draw=black, fill=#1, line width=0.25pt] (0,0) circle (0.7ex);%
}

\newcommand{\clusterCell}[2]{%
  \clusterDot{#1}\hspace{0.45em}#2%
}

\usepackage{newfloat}
\usepackage{listings}
\DeclareCaptionStyle{ruled}{labelfont=normalfont,labelsep=colon,strut=off} 
\floatstyle{ruled}
\newfloat{listing}{tb}{lst}{}
\floatname{listing}{Listing}

\usepackage{booktabs}

\title{Diffusing in the Right Space:\\
A Systematic Study of Latent Diffusability}

\author{
Tianxiong Zhong,
Xingye Tian\footnotemark[2],
Xuebo Wang,
Xin Tao,
Pengfei Wan
}

\affiliations{
Kling Team, Kuaishou Technology\\
inkosizhong@gmail.com,
\{tianxingye,wangxuebo,taoxin,wanpengfei\}@kuaishou.com\\
\footnotemark[2]Corresponding Author
\raisebox{-0.15em}{\includegraphics[height=0.9em]{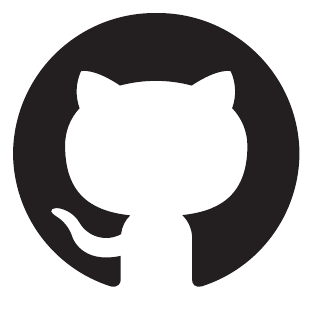}}~
\url{https://github.com/KlingAIResearch/diffusing-right-space}
}

\begin{document}

\nocopyright
\twocolumn[{%
\renewcommand\twocolumn[1][]{#1}%
\maketitle
\vspace{-0.6cm}
\begin{center}
\captionsetup{type=figure}
\centering
\includegraphics[width=\linewidth]{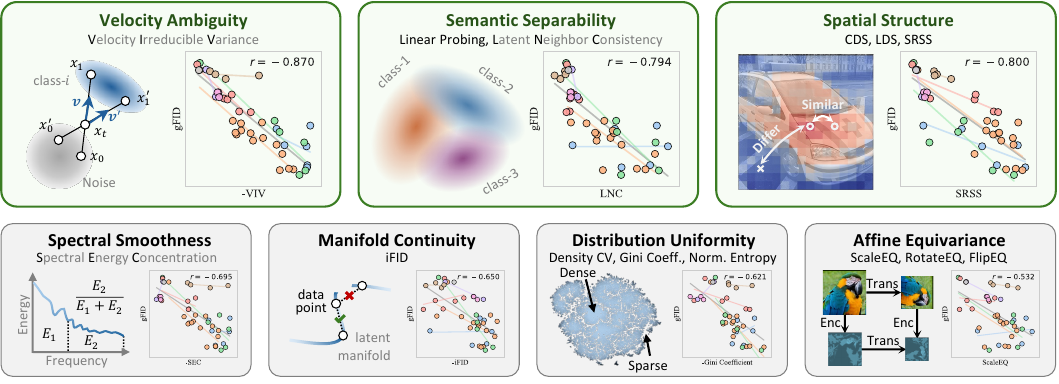}
\caption{Different perspectives for observing latent properties. Each scatter corresponds to a tokenizer with different latent properties. Scatters with same color belong to the same regularization method.}
\label{fig:teaser}
\end{center}%
}]

\input{sections/0_abstract}
\input{sections/1_introduction}
\input{sections/2_metrics}
\input{sections/3_experiment}
\input{sections/4_related_works}
\input{sections/5_conclusion}

\bibliography{aaai2027}


\input{sections/X_appendix}

\end{document}

%% file: sections/0_abstract.tex
\begin{abstract}
Latent diffusion models leverage visual tokenizers to compress images into latent spaces for efficient generative modeling.
However, better reconstruction quality of a tokenizer does not necessarily translate into better generation quality, suggesting that latent representations should be evaluated not only by fidelity but also by their diffusability.
Recent studies have proposed diverse explanations for diffusion-friendly latent spaces, including semantic separability, affine equivariance, distribution uniformity, spatial structure, spectral smoothness, and manifold continuity.
Yet these properties are often validated on a limited set of tokenizers, leaving it unclear which factors are most predictive of downstream generation quality and whether such conclusions hold beyond the specific settings in which they are introduced.
In this work, we conduct a systematic study of latent diffusability by training a large collection of tokenizers with diverse regularization strategies, architectures, and latent configurations, and evaluating them with multiple downstream diffusion backbones.
Our analysis identifies several latent properties that consistently correlate with generation quality and exhibit strong generalization across experimental settings.
Beyond existing metrics, we introduce Velocity Irreducible Variance (VIV), a measure of velocity ambiguity induced by trajectory crossings.
Extensive experiments show that VIV is one of the most stable predictors of generation quality.
\end{abstract}

%% file: sections/1_introduction.tex
\section{Introduction}
The success of latent diffusion models~\cite{rombach2022ldm,labs2025flux,wu2025qwen-image,li2024hunyuan,esser2024sd3} depends not only on the capacity of the diffusion backbone, but also critically on the property of the latent space produced by the tokenizer.
A tokenizer with better reconstruction quality does not necessarily lead to better generation quality, revealing a fundamental mismatch between pixel-level compression and diffusion-friendly representation learning.
This raises a central question: \textit{what kind of latent space is easier for diffusion models to learn?}.

Recent studies have proposed diverse explanations for latent diffusability, including semantic separability~\cite{yao2025VTP,zheng2025rae}, affine equivariance~\cite{kouzelis2025eqvae}, distribution uniformity~\cite{yao2025vavae}, spatial structure~\cite{singh2025irepa}, spectral smoothness~\cite{skorokhodov2025se,fan2025uae}, and manifold continuity~\cite{xu2026ifid}.
However, these properties are offen validated on a limited set of tokenizers.
Moreover, each study typically introduces a particular regularization strategy together with a proxy metric that explains its own improvement.
As a result, it remains unclear \textit{which latent properties are truly predictive of downstream generation quality}, and \textit{whether such conclusions generalize beyond the specific settings in which they are introduced}.

To answer these questions, we conduct a systematic study of latent diffusability.
We construct a large-scale evaluation covering diverse tokenizers trained with different latent regularization strategies~\cite{yao2025vavae,yu2024repa,kouzelis2025eqvae,liu2025ssvae}, tokenizer architectures, and latent configurations.
For each tokenizer, we train multiple downstream diffusion models with different backbones and capacities, enabling a controlled correlation analysis between latent-space properties and generation quality.
This design allows us to compare existing perspectives under a unified evaluation protocol.

To complement existing perspectives, we introduce Velocity Irreducible Variance (VIV), a measure of velocity ambiguity induced by trajectory crossings.
In Flow Matching~\cite{liu2022rectifiedflow}, multiple source-target pairs may induce different velocities at the same interpolated state, leading to an irreducible component in the velocity prediction objective.
We model the class-conditional latent distribution as an anisotropic Gaussian, and show that VIV admits an analytic form determined by the principal standard deviations of the within-class covariance.
This analysis suggests that intra-class compactness and spectral anisotropy are beneficial for reducing the ambiguity.

Our empirical analysis reveals that semantic separability, spatial structure, and VIV consistently exhibit strong correlations with generation quality across different diffusion backbones and tokenizer settings.
Beyond single-perspective analysis, we further conduct a dual-perspective joint analysis and find that a linear model using semantic separability and spatial structure as predictors explains gFID better than either factor alone.
These results suggest that latent diffusability is a multi-faceted property.

Our contributions are summarized as follows:
\begin{itemize}
\item We provide a systematic study of latent diffusability by evaluating diverse latent-space properties across tokenizer architectures, latent configurations, and downstream diffusion backbones.
\item We propose VIV, a flow-based metric that quantifies velocity ambiguity in Flow Matching.
\item We identify VIV, semantic separability, and spatial structure as consistently effective predictors of downstream generation quality across diverse experimental settings.
\end{itemize}

%% file: sections/2_metrics.tex
\section{Perspectives and Metrics}
We focus on the diffusability of latent spaces under controlled settings, where tokenizers have comparable reconstruction quality.
As illustrated in Figure~\ref{fig:teaser}, we summarize seven perspectives for characterizing latent space properties.
We begin with the velocity-based perspective proposed in this paper and describe the computation of the corresponding metric. 
We then briefly review existing perspectives, including semantic separability~\cite{yao2025VTP,chen2025maetok}, spatial structure~\cite{singh2025irepa}, latent smoothness~\cite{skorokhodov2025se,liu2025ssvae}, manifold continuity~\cite{xu2026ifid}, latent uniformity~\cite{yao2025vavae}, and affine equivariance~\cite{kouzelis2025eqvae,skorokhodov2025se}.

\subsection{Velocity Ambiguity}
In the Flow Matching framework, noise $x_0$ and data point $x_1$ are independently sampled from the source and target distributions, respectively, and interpolated at a random time $t$ to obtain $x_t=t\cdot x_1+(1-t)\cdot x_0$.
Diffusion models $\theta$ often predict velocity $v=x_1-x_0$ based on given $x_t$, $t$, and conditional information $y$.
The training objective can be written as follows:
\begin{equation}
\mathcal L(\theta)=\mathbb E_{x_0,x_1,t,y}\left[\|v - v_\theta(x_t,t,y)\|_2^2\right],
\end{equation}
where \(v=x_1-x_0\).
For a fixed interpolated state $x_t$, multiple source-target pairs may induce different velocities~\cite{liu2022rectifiedflow}, leading to an inherent ambiguity.
We hypothesize that the magnitude of this velocity ambiguity affects the diffusability.

Let $v^\star:=v^\star(x_t,t,y)=\mathbb E[v\mid x_t,t,y]$ denote the Bayes-optimal velocity field. 
Then the objective $\mathcal L(\theta)$ can be decomposed into the
following form:
\begin{equation}
{
\color{gray}
\underbrace{
\color{black}
\mathbb E
\left[
\|v^\star-v_\theta(x_t,t,y)\|_2^2
\right]
}_{\substack{\text{Reducible Error}}}
}
+
\underbrace{
\mathbb E
\left[
\|v-v^\star\|_2^2
\right]
}_{\substack{\text{Irreducible Variance}}},
\end{equation}
where the irreducible variance reflects the degree of ambiguity of velocities.
We model the latent distribution of each category with a Gaussian distribution, resulting in a $L$-component Gaussian mixture model (GMM) for the marginal latent distribution, where $L$ denotes the number of categories.
However, the latent representation lies in a high-dimensional space with dimension $d=H\times W\times C$, making direct estimation of the full covariance matrix unreliable when only a limited number of samples $M$ is available, i.e., $d\gg M$.
To address this issue, we adopt the Kronecker Flip-Flop covariance decomposition, which assumes a separable covariance structure between the channel dimension $C$ and the spatial dimension $H\times W$.
Specifically, the full covariance matrix is approximated as:
\begin{equation}
\Sigma \approx \Sigma_s \otimes \Sigma_c,
\quad
\Sigma_c\in\mathbb R^{C\times C},\ 
\Sigma_s\in\mathbb R^{HW\times HW}.
\end{equation}
This assumption reduces the number of covariance parameters to be estimated and increases the effective number of samples for fitting each covariance factor.
For example, when estimating the covariance matrix along the channel dimension, each latent representation can be treated as providing $H\times W$ spatial observations.

For class-conditional generation with a fixed label $y=k$, the target distribution reduces to a single Gaussian, $x_1 \mid y=k \sim \mathcal N(\mu_k,\Sigma_k)$.
Assuming the standard Gaussian source distribution $x_0\sim\mathcal N(0,I)$, the irreducible variance admits an analytic form.
Let $\{\lambda_{k,i}\}_{i=1}^d$ be the eigenvalues of $\Sigma_k$.
At time $t$, the class-wise irreducible variance is given by
\begin{equation}
\mathcal I_k(t)
=
\sum_{i=1}^{d}
\frac{\lambda_{k,i}}
{(1-t)^2+t^2\lambda_{k,i}}.
\label{eq:term2_per_step}
\end{equation}
When $t\sim U(0,1)$, integrating over time yields
\begin{equation}
\mathcal I_k
=
\int_0^1 \mathcal I_k(t)\,\mathrm{d}t
=
\frac{\pi}{2}
\sum_{i=1}^{d}\sqrt{\lambda_{k,i}}.
\label{eq:term2}
\end{equation}
Let $\tau_k:=\mathrm{tr}(\Sigma_k)$ denote the total variance, and $\mathcal A_k :=\mathrm{Var}(\sqrt{\lambda_{k,i}})$ represent the anisotropy of standard-deviation spectrum, Equation~\ref{eq:term2} can be re-written into:

\begin{equation}
\mathcal I_k
=
\frac{\pi}{2}
\sqrt{
d\left(
\tau_k-d\cdot \mathcal A_k)
\right)
},
\quad
\frac{\partial \mathcal I_k}{\partial \tau_k}>0,
\quad
\frac{\partial \mathcal I_k}{\partial \mathcal A_k}<0.
\end{equation}
This analytic form reveals two direct implications for diffusion-friendly latent distributions.

\begin{insightbox}[Insight 1: Intra-class Compactness]
For a fixed spectral shape, reducing the total variance $\tau_k$ shrinks the
average intra-class spread and decreases $\mathcal I_k$.
\end{insightbox}

\begin{insightbox}[Insight 2: Spectral Anisotropy]
When the total variance is controlled, a more anisotropic standard-deviation spectrum, a larger \(\mathcal A_k\), reduces \(\mathcal I_k\).
\end{insightbox}

The overall irreducible variance $\mathcal I$ is obtained by averaging $\mathcal I_k$ over all categories.
For more general settings, such as text-guided generation, the target latent distribution can no longer be reduced to a single class-conditional Gaussian. 
Instead, $x_1$ is sampled from the marginal latent distribution, which is approximated by the GMM. 
Consequently, the marginal distribution of $x_t$ is also a mixture distribution, and $\mathcal I$ can be directly estimated via Monte Carlo sampling.

\subsection{Semantic Separability}
Semantic separability characterizes how well latent representations are organized according to class semantics, reflecting both intra-class compactness and inter-class separation. 
Linear probing~\cite{yu2024repa,yao2025VTP,chen2025maetok} is a widely used evaluation method, which trains a linear classification head
on extracted latents.

\begin{figure}[t]
    \centering
    \includegraphics[width=\linewidth]{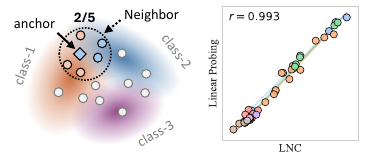}
    \caption{Left: LNC calculates the proportion of samples of the same category within the latent neighborhood. Right: LNC has a high linear correlation with Linear Probing.}
    \label{fig:lnc}
\end{figure}

However, linear probing requires feature extraction over the training set and additional classifier training, making the evaluation computationally expensive.
We therefore introduce Latent Neighbor Consistency (LNC), a validation-set-only proxy for semantic separability.
As shown in Figure~\ref{fig:lnc}, LNC computes the fraction of each latent representation's $K$-nearest neighbors that share the same class label. 
To make the measurement more focused on semantic content, we use pre-computed foreground masks and aggregate only foreground latent pixels.
We observe a strong linear correlation between LNC and linear probing, and thus adopt LNC as an efficient alternative in our analysis.

\subsection{Spatial Structure}
iREPA~\cite{singh2025irepa} studies how the spatial structure of foundation-model representations affects the generation quality of diffusion models under representation alignment~\cite{yu2024repa}. 
Following this line of analysis, we consider three metrics proposed in iREPA: LDS, CDS, and SRSS.
LDS measures whether nearby latent pixels are more similar than distant ones, and CDS quantifies the decay rate of similarity with respect to spatial distance. 
SRSS uses foreground masks to assess whether intra-foreground representations are more consistent than foreground-background representations.
We exclude RMSC because it mainly characterizes the diversity of spatial representations.

\subsection{Latent Smoothness}
Recent analyses of diffusion learning dynamics suggest that high-variance spectral modes are learned faster than low-variance modes, implying that coarse or low-frequency information are typically captured earlier than fine high-frequency details~\cite{wang2026analytical}.
This means that a smaller proportion of high-frequency energy~\cite{skorokhodov2025se,liu2025ssvae,fan2025uae} in the latent space may result in better diffusability.
To quantify this property, we propose a metric Spectral Energy Concentration (SEC), which measures the proportion of spectral energy concentrated in the high-frequency region.

Given a set of latent representations $\mathcal Z=\{z_n\}_{n=1}^{N}$, where $z_n\in\mathbb R^{C\times H\times W}$, we apply the 2D discrete cosine transform (DCT) to each channel independently:
\begin{equation}
\hat z_n = \mathrm{DCT_{2D}}(z_n).
\end{equation}
The average spectral energy at frequency coordinate $(u,v)$ is computed as:
\begin{equation}
E_{u,v}
=
\frac{1}{NC}
\sum_{n=1}^{N}
\left\|
\hat z_{n,:,u,v}
\right\|_2^2.
\end{equation}
Since the low-frequency components of DCT are located near the upper-left corner, we use the Manhattan distance $d(u,v)=u+v$, where a larger value indicates a higher spatial frequency.
Given a threshold ratio $\rho\in[0,1]$, the corresponding frequency threshold is $\tau_\rho = \rho \cdot d(H-1,W-1)$.
Then SEC is defined as the proportion of energy lying outside the low-frequency region:
\begin{equation}
\text{SEC}_\rho=
\frac{
\sum_{u=0}^{H-1}
\sum_{v=0}^{W-1}
\mathbf 1[d(u,v)>\tau_\rho] E_{u,v}
}{
\sum_{u=0}^{H-1}
\sum_{v=0}^{W-1}
E_{u,v}
}.
\end{equation}
A larger SEC indicates that more spectral energy is concentrated in high-frequency components, suggesting a less smooth latent representation.

\subsection{Manifold Continuity}
iFID~\cite{xu2026ifid} and VE~\cite{li2026ve} suggest that the connectivity of latent distributions is closely related to generation quality. 
A continuous latent space is expected to preserve meaningful image semantics and visual realism along local interpolation paths.
Specifically, for each latent representation, iFID first identifies its nearest neighbor in the latent space and then constructs interpolated latents between the two representations. 
These interpolated latents are decoded back into the image space, and the distribution of the decoded images is compared with the
real image distribution using FID. 
A lower iFID indicates that interpolated latents remain closer to the image manifold, suggesting better manifold continuity.

\subsection{Latent Uniformity}
VAVAE~\cite{yao2025vavae} studies latent-space uniformity from the perspective of representation utilization.
A more uniformly utilized latent space can alleviate the concentration of representations in a small number of regions, thereby providing a more regular target distribution for diffusion modeling.
Following VAVAE, we directly adopt its uniformity evaluation protocol.
Specifically, we first extract latent representations from the validation set and project them into a two-dimensional space using t-SNE~\cite{van2008tsne}.
Then, we estimate the density distribution of the projected latent points and compute three statistics to characterize its uniformity: density coefficient of variation, Gini coefficient, and normalized entropy.
A lower density coefficient of variation and Gini coefficient indicate a more even density distribution, while a higher normalized entropy indicates better latent-space uniformity.

\subsection{Affine Equivariance}
Affine Equivariance~\cite{kouzelis2025eqvae,skorokhodov2025se} evaluates whether the tokenizer preserves the geometric transformation structure of the input image.
Such equivariance may provide a more regulated latent representation and may help the downstream diffusion model learn spatial variations more effectively.
Given an input image $x$, we evaluate affine equivariance by comparing the two operator orders, $\mathrm{Enc}\circ\mathrm{Trans}$ and $\mathrm{Trans}\circ\mathrm{Enc}$.
A smaller discrepancy indicates better equivariance.
In our evaluation, we consider two types of transformations: Rotate and Scale.
A higher consistency indicates that the encoder better preserves affine equivariance in the latent space.

%% file: sections/3_experiment.tex
\section{Experiments}
\begin{figure}[t]
    \centering
    \includegraphics[width=\linewidth]{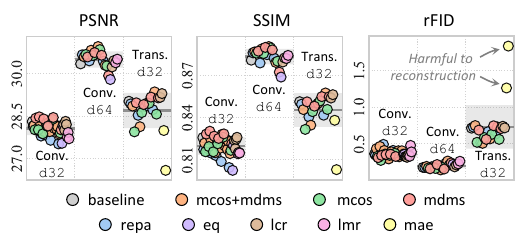}
    \caption{Tokenizers with same architecture and latent configuration have similar reconstruction quality.}
    \label{plot:recon}
\end{figure}
\subsection{Setups}
We trained a serials of tokenizers based on the latent regularization method proposed in existing works~\cite{yao2025vavae,yu2024repa,liu2025ssvae,kouzelis2025eqvae}.
For different regularization methods, we can construct a cluster of tokenizers by adjusting the relevant parameters.
For example, we used various visual foundation models~\cite{oquab2023dinov2,simeoni2025dinov3,radford2021clip,he2022mae,fan2025webssl,chen2021mocov3,bolya2026pe,heinrich2025cradio} for the representation alignment methods.
All tokenizer are trained for 16 epochs on ImageNet~\cite{deng2009imagenet} dataset.

To study whether the conclusions generalize across tokenizer architectures and latent configurations, we evaluate three tokenizer families: 43 convolutional tokenizers with the \texttt{f16d32} latent configuration (\texttt{conv-f16d32}); 22 convolutional tokenizers with the \texttt{f16d64} latent configuration (\texttt{conv-f16d64}); and 21 transformer-based tokenizers with the \texttt{f16d32} latent configuration (\texttt{trans-f16d32}).
As shown in Figure~\ref{plot:recon}, tokenizers within each family have comparable reconstruction quality, ensuring that downstream generative performance is not primarily bounded by reconstruction fidelity.
The proxy metrics are computed on either the validation set or its masked variant~\cite{gao2022s919}.
For each tokenizer, we train different diffusion models: SiT-B, SiT-XL, LightningDiT-B, and LightningDiT-XL.
The training strategy follows the official configuration.
We train 400k steps for SiT-B~\cite{ma2024sit}, 80k steps for SiT-XL, and 100k steps for LightningDiT~\cite{yao2025vavae} models~\cite{yao2025vavae}, respectively.
In this section, we use gFID~\cite{heusel2017fid} to represent the generation quality, and we also provide the results for IS~\cite{salimans2016IS} and FDr$^6$~\cite{yang2026fdloss} in the appendix.

\begin{figure*}[t]
    \centering
    \includegraphics[width=\linewidth]{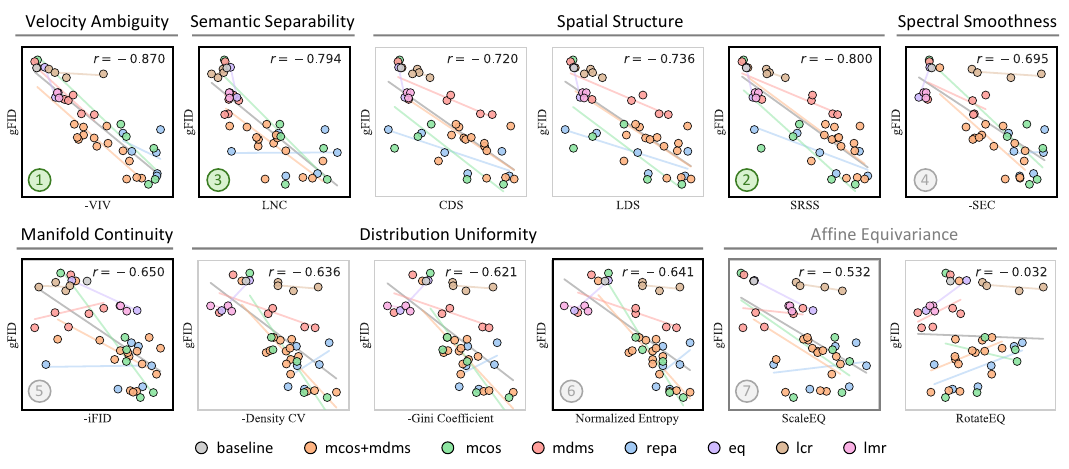}
    \caption{Correlation between different perspectives and generation quality on \texttt{conv-f16d32} and SiT-B. The most relevant metric for each perspective is highlighted with bold border. The order of relevance is given by number.}
    \label{plot:sit_b}
\end{figure*}

\begin{figure*}[!t]
    \centering
    \includegraphics[width=\linewidth]{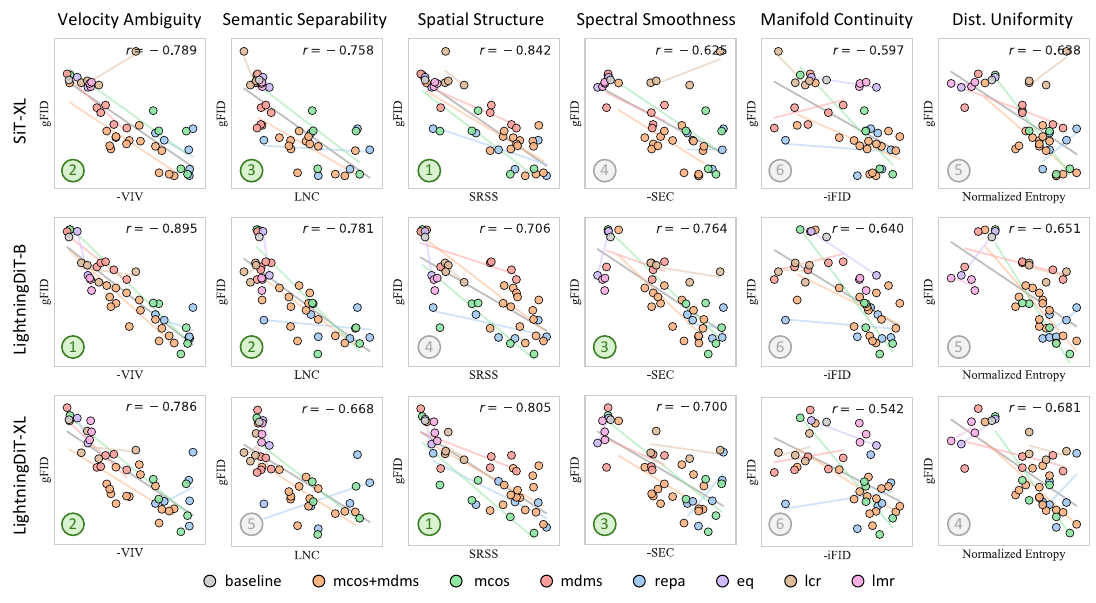}
    \caption{Correlation analysis on \texttt{conv-f16d32} across various downstream diffusion backbones.}
    \label{plot:other-dit}
\end{figure*}

\begin{figure*}[!t]
    \centering
    \includegraphics[width=\linewidth]{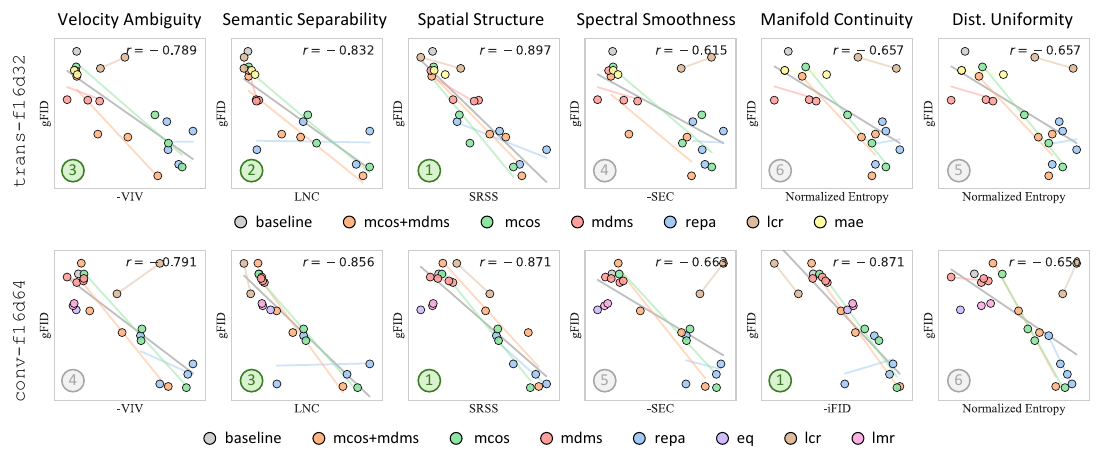}
    \caption{Correlation analysis on SiT-B across various tokenizer families.}
    \label{plot:other-tok}
\end{figure*}
\begin{figure*}[!t]
    \centering
    \includegraphics[width=\linewidth]{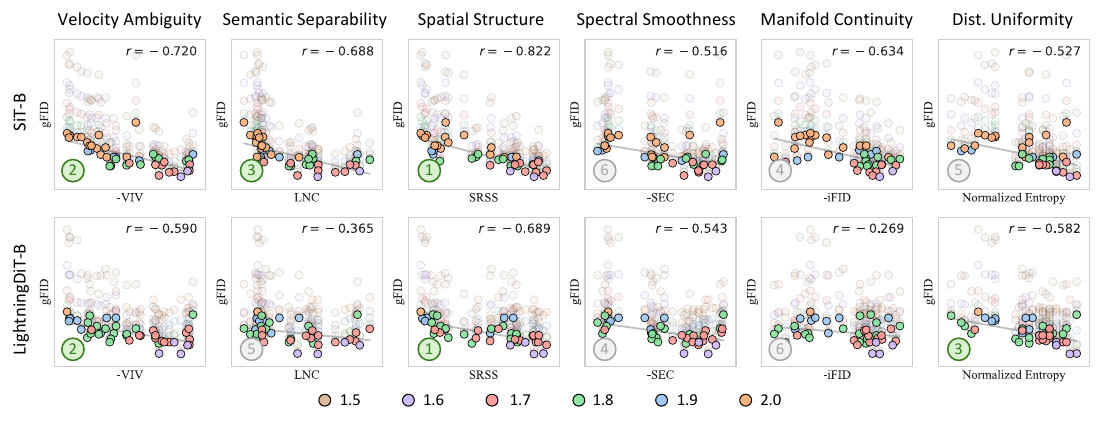}
    \caption{Impact of classifier-Free guidance on \texttt{conv-f16d32}. The optimal CFG for each latent space is highlighted.}
    \label{plot:cfg}
\end{figure*}

\subsection{Which Perspective Matters?}
As shown in Figure~\ref{plot:sit_b}, we enumerate the relationships between different proxy metrics and generation quality from each perspective.
The metric with the highest relevance within each perspective is highlighted, and it is used as the main proxy in subsequent experiments.
We ranked the perspectives based on relevance, with Velocity Ambiguity, Semantic Separability, and Spatial Structure standing out.
The Pearson coefficient for VIV and gFID reached 0.87.
In contrast, the correlations among Manifold Continuity, Distribution Uniformity, and Affine Equivariance are relatively low, and the trends within each regularized cluster differ significantly.
In particular, since the Affine Equivariance has the lowest correlation and the two metrics lack consistency, we ignored this perspective in subsequent analyses, and the corresponding results are presented in the appendix.

\subsection{Generalization across Diffusion Backbones}
Figure~\ref{plot:other-dit} exhibits the results among SiT-XL, LightningDiT-B, and LightningDiT-XL.
Among the four Diffusion Models, Velocity Ambiguity and Spatial Structure are the most stable, while Semantic Separability and Spectral Smoothness are relatively better.
It is worth noting that as the diffusion capacity increases from B to XL, SRSS fits better, while the correlation of other metrics decreases or remained unchanged.
SiT and LightningDiT also show differences in property preferences.
For example, LNS performs better on SiT, while SEC performs better on LightningDiT.
We believe this difference mainly stems from the different timestep sampling strategies.
(Uniform for SiT and LogNorm for LightningDIT).

\subsection{Generalization across Tokenizer Families}
In Figure~\ref{plot:other-tok}, we further evaluate whether the properties generalize across different tokenizer families on SiT-B.
Across the families, Velocity Ambiguity, Semantic Separability, and Spatial Structure remain effective.
We also observe that iFID~\cite{xu2026ifid} shows a particularly high correlation on the \texttt{conv-f16d64} family, achieving performance comparable to SRSS.
However, iFID is less stable in our overall experiments.
We hypothesize that this is because we intentionally control the reconstruction quality of tokenizers within the same family to be similar.
Under this setting, reconstruction-oriented metrics have a relatively limited dynamic range, making them less reliable for explaining the remaining differences in downstream generation quality.

\begin{figure*}[!t]
    \centering
    \includegraphics[width=\linewidth]{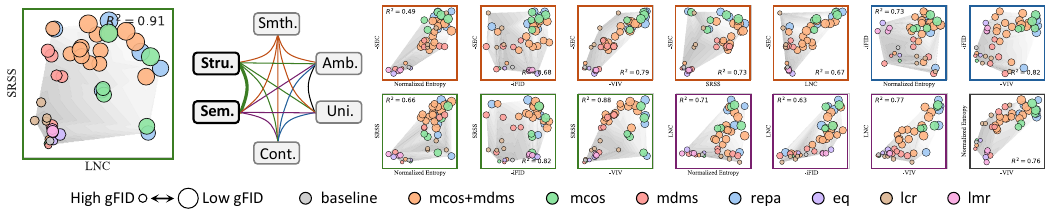}
    \caption{Dual-perspective regression of gFID on \texttt{conv-f16d32}, where the size of the bubble corresponds to the gFID, and the terrain of the background represents the trend. Border colors facilitate quick checking of perspective combinations.}
    \label{plot:bubble}
\end{figure*}
\subsection{Impact of Classifier-Free Guidance}
We evaluate the w/ CFG results on SiT-B and LightningDit-B, varying CFG scale~\cite{ho2022cfg} from 1.0 to 3.0.
As shown in Figure~\ref{plot:cfg}, we present the results in the range of 1.5 to 2.0, because we find that the optimal CFG for all experiments lies in this range (see Appendix).
Each tokenizer corresponds to a vertical column of scatter, where the optimal gFID configuration is highlighted.
Experimental results show that Velocity Ambiguity and Spatial Structure still provide the best and most stable fit.
We also find that the configuration with CFG seems to further amplify the framework differences in the Diffusion backbones.

\subsection{Complementarity across Perspectives}
As illustrated in Figure~\ref{plot:bubble}, we enumerated combinations of two perspectives to regress gFID.
The two axes in the figure correspond to the proxy metrics, and the size of the bubble reflects gFID.
First, most of the perspectives are approximately orthogonal to each other, which allows them to open up a large area on a two-dimensional plane.
Only several perspectives are found a weak correlation.
For example, Spectral Smoothness, Distribution Uniformity, and Velocity Ambiguity exhibit a certain collinearity.
This collinearity may stem from correlations in the underlying mechanisms, but may also originate from the way the tokenizers are constructed.
We will consider more tokenizers and conduct further research on this phenomenon.
On the other hand, we found that the space spanned by SRSS and LNC can fit gFID with an $R^2=0.91$, indicating that scatters located at Pareto optimality in terms of Spatial Structure and Semantic Separability will have better generation quality.
This suggests that a comprehensive evaluation of latent space from multiple perspectives may be more accurate and reliable.

\begin{figure}[t]
    \centering
    \includegraphics[width=\linewidth]{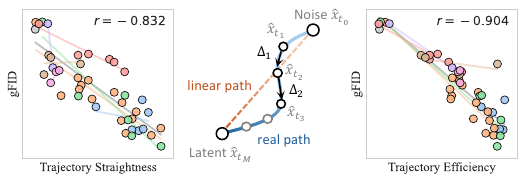}
    \caption{Latent spaces with better generation quality tend to produce straighter and more efficient trajectories.}
    \label{plot:path}
\end{figure}

\subsection{Better Latents Induce More Efficient Transport}
We further find that latent spaces with better generation quality tend to induce simpler learned velocity fields, reflected by straighter ODE trajectories.
This provides a post-hoc view of how latent-space properties may affect the dynamics learned by diffusion models.
Specifically, we record the full denoising trajectory $\{\hat{x}_{t_i}\}_{i=0}^{M}$ of the trained diffusion model, where $\hat{x}_{t_0}$ is the initial Gaussian noise and $\hat{x}_{t_M}$ is the generated latent.
For each segment, we define $\Delta_i=\hat{x}_{t_{i+1}}-\hat{x}_{t_i}$.
We measure the local straightness of the trajectory by the average cosine similarity between adjacent segments:
\begin{equation}
\mathrm{Straightness}
=
\frac{1}{M-1}
\sum_{i=0}^{M-2}
\frac{\langle \Delta_i, \Delta_{i+1}\rangle}
{\|\Delta_i\|_2\|\Delta_{i+1}\|_2}.
\end{equation}
We also measure the global efficiency by comparing the endpoint displacement with the accumulated path length:
\begin{equation}
\mathrm{Efficiency}
=
\frac{\|\hat{x}_{t_M}-\hat{x}_{t_0}\|_2}
{\sum_{i=0}^{M-1}\|\Delta_i\|_2}.
\end{equation}
As shown in Figure~\ref{plot:path}, both metrics are highly correlated with gFID.
This suggests that better latent spaces lead the diffusion model to follow more direct and less redundant ODE paths.
This observation indicates that latent-space properties may influence the complexity of the target velocity field, or equivalently the difficulty of fitting the learned dynamics.

\begin{figure}[t]
    \centering
    \includegraphics[width=\linewidth]{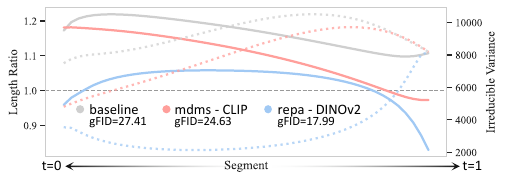}
    \caption{Per-segment length ratio along ODE trajectories (solid), and estimated irreducible variance (dotted).}
    \label{plot:seg_curve}
\end{figure}

Figure~\ref{plot:seg_curve} further visualizes the per-segment length ratio
$M\cdot \|\Delta_i\|_2/\|\hat{x}_{t_M}-\hat{x}_{t_0}\|_2$
along the ODE trajectory for three representative tokenizers with poor, medium, and strong generation quality.
A ratio of 1 corresponds to the segment length of the linear path, while ratios above or below 1 indicate more aggressive or more conservative updates, respectively.
We observe that better latent spaces keep the length ratio closer to 1, suggesting that the learned ODE trajectory follows a more balanced and efficient transport schedule.
In contrast, the baseline deviates more significantly from the linear-path schedule, especially in the early and middle denoising stages.

We also overlay the irreducible variance estimated by Equation~\ref{eq:term2_per_step}.
The irreducible variance and the learned length ratio exhibit a highly consistent but opposite pattern, where regions with larger irreducible variance tend to correspond to smaller learned step lengths.
In regions with higher velocity $v$ ambiguity, the Bayes-optimal velocity $v^\star$ tends to have a smaller magnitude. 
Since $v_\theta$ is trained to approximate this Bayes-optimal velocity, it naturally exhibits reduced magnitudes in these regions.

%% file: sections/4_related_works.tex
\section{Related Work}
\noindent\textbf{Analysis Paradigm.}
iREPA~\cite{singh2025irepa} studies representation alignment in diffusion training and investigates whether global semantic information or spatial structure of the target representation matters more.
We extend this analytical paradigm to the properties of latent space.

\noindent\textbf{Broader Tokenizer Representations.}
Recent works, like DC-AE 1.5~\cite{chen2025dcae1.5}, RAE~\cite{zheng2025rae}, and DM-VAE~\cite{ye2025dmvae}, introduce different architectures, regularization strategies, or representation priors for visual tokenization.
Meanwhile, 1D tokenizers~\cite{yu2024titok,bachmann2025flextok,chen2025maetok} represent images as sequential tokens, providing another form of latent representation for generative modeling.
Our analysis framework can be extended to these representations to further study whether the identified latent properties remain predictive across broader tokenizer families.
Lastly, we primarily compare tokenizers under the same architecture, latent configuration, and comparable reconstruction quality, while leaving cross-family comparisons for future work.

%% file: sections/5_conclusion.tex
\section{Conclusion}
In this work, we present a systematic study of latent diffusability, aiming to understand what makes a latent space easier for diffusion models to learn.
Instead of focusing on a single tokenizer design or regularization strategy, we evaluate diverse latent-space properties across different tokenizer architectures, latent configurations, and downstream diffusion backbones.
Our analysis shows that diffusion-friendly latent spaces are jointly shaped by semantic, structural, and spectral properties.
To provide a complementary perspective, we introduce Velocity Irreducible Variance (VIV), which quantifies the intrinsic velocity ambiguity in Flow Matching.
By modeling class-conditional latent distributions with anisotropic Gaussians, VIV connects downstream learnability to intra-class compactness and spectral anisotropy.
Empirically, VIV exhibits stable correlations with generation quality across a wide range of settings.
Overall, our findings suggest that latent diffusability should be understood as a multi-faceted property rather than a consequence of any single regularization objective.

%% file: sections/X_appendix.tex
\setcounter{page}{1}
\onecolumn{%
\renewcommand\onecolumn[1][]{#1}%
\begin{center}
    \maketitleappendix
    \centering
    \begin{minipage}{\textwidth}
    \centering
    \small
    \captionof{table}{
        Summary of all tokenizers, including identifier, architecture, latent configuration, cluster, and variant.
        For alignment-based clusters
        \clusterDot{cRepa} \clusterDot{cVF} \clusterDot{cMcos} \clusterDot{cMdms},
        the variants specify the foundation models used for alignment.
        For eq \clusterDot{cEq}, the variants specify the transformation operators.
        For lcr \clusterDot{cLcr},
        \texttt{w} and \texttt{th} denote the loss weight and threshold.
        For lmr \clusterDot{cLmr},
        \texttt{p}$a$-$b$-$c$ denotes the probabilities of masking 25\%, 50\%, and 75\% of tokens.
        For mae \clusterDot{cMae},
        \texttt{r} denotes the maximum masking ratio.
    }
    \label{tab:tokenizer}
    \setlength{\tabcolsep}{5pt}
    \begin{tabular}{ccclcccclc}
    \toprule
    \textbf{ID} & \textbf{Arch.} & \textbf{Config.} & \textbf{Cluster} & \textbf{Variant} & \textbf{ID} & \textbf{Arch.} & \textbf{Config.} & \textbf{Cluster} & \textbf{Variant} \\
    \midrule
    1 & Conv. & \texttt{f16d32} & \clusterCell{cBaseline}{baseline} & - & 44 & Conv. & \texttt{f16d64} & \clusterCell{cBaseline}{baseline} & - \\
    2 & Conv. & \texttt{f16d32} & \clusterCell{cRepa}{repa} & \texttt{CLIP-L} & 45 & Conv. & \texttt{f16d64} & \clusterCell{cRepa}{repa} & \texttt{CLIP-L} \\
    3 & Conv. & \texttt{f16d32} & \clusterCell{cVF}{mcos+mdms} & \texttt{CLIP-L} & 46 & Conv. & \texttt{f16d64} & \clusterCell{cVF}{mcos+mdms} & \texttt{CLIP-L} \\
    4 & Conv. & \texttt{f16d32} & \clusterCell{cMcos}{mcos} & \texttt{CLIP-L} & 47 & Conv. & \texttt{f16d64} & \clusterCell{cMcos}{mcos} & \texttt{CLIP-L} \\
    5 & Conv. & \texttt{f16d32} & \clusterCell{cMdms}{mdms} & \texttt{CLIP-L} & 48 & Conv. & \texttt{f16d64} & \clusterCell{cMdms}{mdms} & \texttt{CLIP-L} \\
    6 & Conv. & \texttt{f16d32} & \clusterCell{cRepa}{repa} & \texttt{CRadio-B} & 49 & Conv. & \texttt{f16d64} & \clusterCell{cRepa}{repa} & \texttt{DINOv2-B} \\
    7 & Conv. & \texttt{f16d32} & \clusterCell{cVF}{mcos+mdms} & \texttt{CRadio-B} & 50 & Conv. & \texttt{f16d64} & \clusterCell{cVF}{mcos+mdms} & \texttt{DINOv2-B} \\
    8 & Conv. & \texttt{f16d32} & \clusterCell{cMcos}{mcos} & \texttt{CRadio-B} & 51 & Conv. & \texttt{f16d64} & \clusterCell{cMcos}{mcos} & \texttt{DINOv2-B} \\
    9 & Conv. & \texttt{f16d32} & \clusterCell{cMdms}{mdms} & \texttt{CRadio-B} & 52 & Conv. & \texttt{f16d64} & \clusterCell{cMdms}{mdms} & \texttt{DINOv2-B} \\
    10 & Conv. & \texttt{f16d32} & \clusterCell{cVF}{mcos+mdms} & \texttt{CRadio-L} & 53 & Conv. & \texttt{f16d64} & \clusterCell{cRepa}{repa} & \texttt{DINOv3-B} \\
    11 & Conv. & \texttt{f16d32} & \clusterCell{cRepa}{repa} & \texttt{DINOv2-B} & 54 & Conv. & \texttt{f16d64} & \clusterCell{cVF}{mcos+mdms} & \texttt{DINOv3-B} \\
    12 & Conv. & \texttt{f16d32} & \clusterCell{cVF}{mcos+mdms} & \texttt{DINOv2-B} & 55 & Conv. & \texttt{f16d64} & \clusterCell{cMcos}{mcos} & \texttt{DINOv3-B} \\
    13 & Conv. & \texttt{f16d32} & \clusterCell{cMcos}{mcos} & \texttt{DINOv2-B} & 56 & Conv. & \texttt{f16d64} & \clusterCell{cMdms}{mdms} & \texttt{DINOv3-B} \\
    14 & Conv. & \texttt{f16d32} & \clusterCell{cMdms}{mdms} & \texttt{DINOv2-B} & 57 & Conv. & \texttt{f16d64} & \clusterCell{cRepa}{repa} & \texttt{MAE-L} \\
    15 & Conv. & \texttt{f16d32} & \clusterCell{cVF}{mcos+mdms} & \texttt{DINOv2-L} & 58 & Conv. & \texttt{f16d64} & \clusterCell{cVF}{mcos+mdms} & \texttt{MAE-L} \\
    16 & Conv. & \texttt{f16d32} & \clusterCell{cRepa}{repa} & \texttt{DINOv3-B} & 59 & Conv. & \texttt{f16d64} & \clusterCell{cMcos}{mcos} & \texttt{MAE-L} \\
    17 & Conv. & \texttt{f16d32} & \clusterCell{cVF}{mcos+mdms} & \texttt{DINOv3-B} & 60 & Conv. & \texttt{f16d64} & \clusterCell{cMdms}{mdms} & \texttt{MAE-L} \\
    18 & Conv. & \texttt{f16d32} & \clusterCell{cMcos}{mcos} & \texttt{DINOv3-B} & 61 & Conv. & \texttt{f16d64} & \clusterCell{cEq}{eq} & \texttt{scale} \\
    19 & Conv. & \texttt{f16d32} & \clusterCell{cMdms}{mdms} & \texttt{DINOv3-B} & 62 & Conv. & \texttt{f16d64} & \clusterCell{cLcr}{lcr} & \texttt{w0.02-th0.75} \\
    20 & Conv. & \texttt{f16d32} & \clusterCell{cVF}{mcos+mdms} & \texttt{DINOv3-L} & 63 & Conv. & \texttt{f16d64} & \clusterCell{cLcr}{lcr} & \texttt{w0.05-th0.90} \\
    21 & Conv. & \texttt{f16d32} & \clusterCell{cVF}{mcos+mdms} & \texttt{LangPE-L} & 64 & Conv. & \texttt{f16d64} & \clusterCell{cLmr}{lmr} & \texttt{p0.1-0.10-0.10} \\
    22 & Conv. & \texttt{f16d32} & \clusterCell{cRepa}{repa} & \texttt{MAE-L} & 65 & Conv. & \texttt{f16d64} & \clusterCell{cLmr}{lmr} & \texttt{p0.1-0.05-0.05} \\
    23 & Conv. & \texttt{f16d32} & \clusterCell{cVF}{mcos+mdms} & \texttt{MAE-L} & 66 & Trans. & \texttt{f16d32} & \clusterCell{cBaseline}{baseline} & - \\
    24 & Conv. & \texttt{f16d32} & \clusterCell{cMcos}{mcos} & \texttt{MAE-L} & 67 & Trans. & \texttt{f16d32} & \clusterCell{cRepa}{repa} & \texttt{CLIP-L} \\
    25 & Conv. & \texttt{f16d32} & \clusterCell{cMdms}{mdms} & \texttt{MAE-L} & 68 & Trans. & \texttt{f16d32} & \clusterCell{cVF}{mcos+mdms} & \texttt{CLIP-L} \\
    26 & Conv. & \texttt{f16d32} & \clusterCell{cVF}{mcos+mdms} & \texttt{MoCov3-L} & 69 & Trans. & \texttt{f16d32} & \clusterCell{cMcos}{mcos} & \texttt{CLIP-L} \\
    27 & Conv. & \texttt{f16d32} & \clusterCell{cVF}{mcos+mdms} & \texttt{PE-B} & 70 & Trans. & \texttt{f16d32} & \clusterCell{cMdms}{mdms} & \texttt{CLIP-L} \\
    28 & Conv. & \texttt{f16d32} & \clusterCell{cVF}{mcos+mdms} & \texttt{PE-L} & 71 & Trans. & \texttt{f16d32} & \clusterCell{cRepa}{repa} & \texttt{DINOv2-B} \\
    29 & Conv. & \texttt{f16d32} & \clusterCell{cVF}{mcos+mdms} & \texttt{SpatialPE-B} & 72 & Trans. & \texttt{f16d32} & \clusterCell{cVF}{mcos+mdms} & \texttt{DINOv2-B} \\
    30 & Conv. & \texttt{f16d32} & \clusterCell{cVF}{mcos+mdms} & \texttt{SpatialPE-L} & 73 & Trans. & \texttt{f16d32} & \clusterCell{cMcos}{mcos} & \texttt{DINOv2-B} \\
    31 & Conv. & \texttt{f16d32} & \clusterCell{cRepa}{repa} & \texttt{WebSSL-300m} & 74 & Trans. & \texttt{f16d32} & \clusterCell{cMdms}{mdms} & \texttt{DINOv2-B} \\
    32 & Conv. & \texttt{f16d32} & \clusterCell{cVF}{mcos+mdms} & \texttt{WebSSL-300m} & 75 & Trans. & \texttt{f16d32} & \clusterCell{cRepa}{repa} & \texttt{DINOv3-B} \\
    33 & Conv. & \texttt{f16d32} & \clusterCell{cMcos}{mcos} & \texttt{WebSSL-300m} & 76 & Trans. & \texttt{f16d32} & \clusterCell{cVF}{mcos+mdms} & \texttt{DINOv3-B} \\
    34 & Conv. & \texttt{f16d32} & \clusterCell{cMdms}{mdms} & \texttt{WebSSL-300m} & 77 & Trans. & \texttt{f16d32} & \clusterCell{cMcos}{mcos} & \texttt{DINOv3-B} \\
    35 & Conv. & \texttt{f16d32} & \clusterCell{cEq}{eq} & \texttt{scale} & 78 & Trans. & \texttt{f16d32} & \clusterCell{cMdms}{mdms} & \texttt{DINOv3-B} \\
    36 & Conv. & \texttt{f16d32} & \clusterCell{cEq}{eq} & \texttt{flip} & 79 & Trans. & \texttt{f16d32} & \clusterCell{cRepa}{repa} & \texttt{MAE-L} \\
    37 & Conv. & \texttt{f16d32} & \clusterCell{cLcr}{lcr} & \texttt{w0.02-th0.60} & 80 & Trans. & \texttt{f16d32} & \clusterCell{cVF}{mcos+mdms} & \texttt{MAE-L} \\
    38 & Conv. & \texttt{f16d32} & \clusterCell{cLcr}{lcr} & \texttt{w0.02-th0.75} & 81 & Trans. & \texttt{f16d32} & \clusterCell{cMcos}{mcos} & \texttt{MAE-L} \\
    39 & Conv. & \texttt{f16d32} & \clusterCell{cLcr}{lcr} & \texttt{w0.05-th0.70} & 82 & Trans. & \texttt{f16d32} & \clusterCell{cMdms}{mdms} & \texttt{MAE-L} \\
    40 & Conv. & \texttt{f16d32} & \clusterCell{cLcr}{lcr} & \texttt{w0.05-th0.90} & 83 & Trans. & \texttt{f16d32} & \clusterCell{cLcr}{lcr} & \texttt{w0.02-th0.75} \\
    41 & Conv. & \texttt{f16d32} & \clusterCell{cLmr}{lmr} & \texttt{p0.1-0.15-0.15} & 84 & Trans. & \texttt{f16d32} & \clusterCell{cLcr}{lcr} & \texttt{w0.05-th0.90} \\
    42 & Conv. & \texttt{f16d32} & \clusterCell{cLmr}{lmr} & \texttt{p0.1-0.10-0.10} & 85 & Trans. & \texttt{f16d32} & \clusterCell{cMae}{mae} & \texttt{r0.3} \\
    43 & Conv. & \texttt{f16d32} & \clusterCell{cLmr}{lmr} & \texttt{p0.1-0.05-0.05} & 86 & Trans. & \texttt{f16d32} & \clusterCell{cMae}{mae} & \texttt{r0.7} \\
    \bottomrule
    \end{tabular}
    \end{minipage}
\end{center}%
}

\section{Implementation Details}
Table~\ref{tab:tokenizer} enumerates all the tokenizers we evaluated, and the ID and cluster colors in all figures in the appendix are follow this specification.
Specifically, all tokenizers are build upon the Variational Autoencoder approach, and trained with a standard objective~\cite{yao2025vavae}:
\begin{equation}
\mathcal{L}=\mathcal{L}_\text{L1}+\lambda_1\mathcal{L}_\text{LPIPS}+\lambda_2\cdot\lambda_\nabla\mathcal{L}_\text{GAN}+\lambda_3\mathcal{L}_\text{KL},
\end{equation}
where $\lambda_1=1$, $\lambda_2=0.5$, $\lambda_3=10^{-6}$, and $\lambda_\nabla$ represents a gradient-driven adaptive weight.

For the diffusion models, we follow the official implementations~\cite{ma2024sit,yao2025vavae} and enable QKNorm to improve training stability.
To ensure an efficient and fair comparison, we fix 50 sampling steps for all approaches.
The configurations are detailed in Table~\ref{tab:dit}.

\begin{table}[h]
    \centering
    \caption{Detailed configurations for diffusion models.}
    \small
    \begin{tabular}{lcccc}
    \toprule
    & SiT-B & SiT-XL & LightningDiT-B & LightningDiT-XL \\
    \midrule
    \multicolumn{5}{l}{\textit{Architecture}} \\
    $\quad$ patch size & 1 & 1 & 1 & 1 \\
    $\quad$ \#layers & 12 & 28 & 12 & 28 \\
    $\quad$ \#hidden dimension & 768 & 1152 & 768 & 1152 \\
    $\quad$ \#head & 12 & 16 & 12 & 16 \\
    $\quad$ position embedding & Sinusoidal & Sinusoidal & RoPE & RoPE \\
    $\quad$ layer normalization & LayerNorm & LayerNorm & RMSNorm & RMSNorm \\
    $\quad$ feedforward network & MLP & MLP & SwiGLU & SwiGLU \\
    $\quad$ QKNorm & \Checkmark & \Checkmark & \Checkmark & \Checkmark \\
    \midrule
    \multicolumn{5}{l}{\textit{Optimization}} \\
    $\quad$ timestep sampling & Uniform & Uniform & Logit Normal & Logit Normal \\
    $\quad$ loss & MSE & MSE & MSE+Cosine & MSE+Cosine \\
    $\quad$ training steps & 400k & 80k & 100k & 100k \\
    $\quad$ batch size & 256 & 256 & 1024 & 1024 \\
    $\quad$ learning rate & 1e-4 & 1e-4 & 2e-4 & 2e-4 \\
    $\quad$ AdamW $\beta_2$ & 0.999 & 0.999 & 0.95 & 0.95 \\
    \midrule
    \multicolumn{5}{l}{\textit{Sampling}} \\
    $\quad$ mode & ODE & ODE & ODE & ODE \\
    $\quad$ sampler & Euler & Euler & Euler & Euler \\
    $\quad$ steps & 50 & 50 & 50 & 50 \\
    \bottomrule
    \end{tabular}
    \label{tab:dit}
\end{table}

\clearpage
\section{Detailed Figures for gFID}
\begin{figure}[!h]
    \centering
    \includegraphics[width=\linewidth]{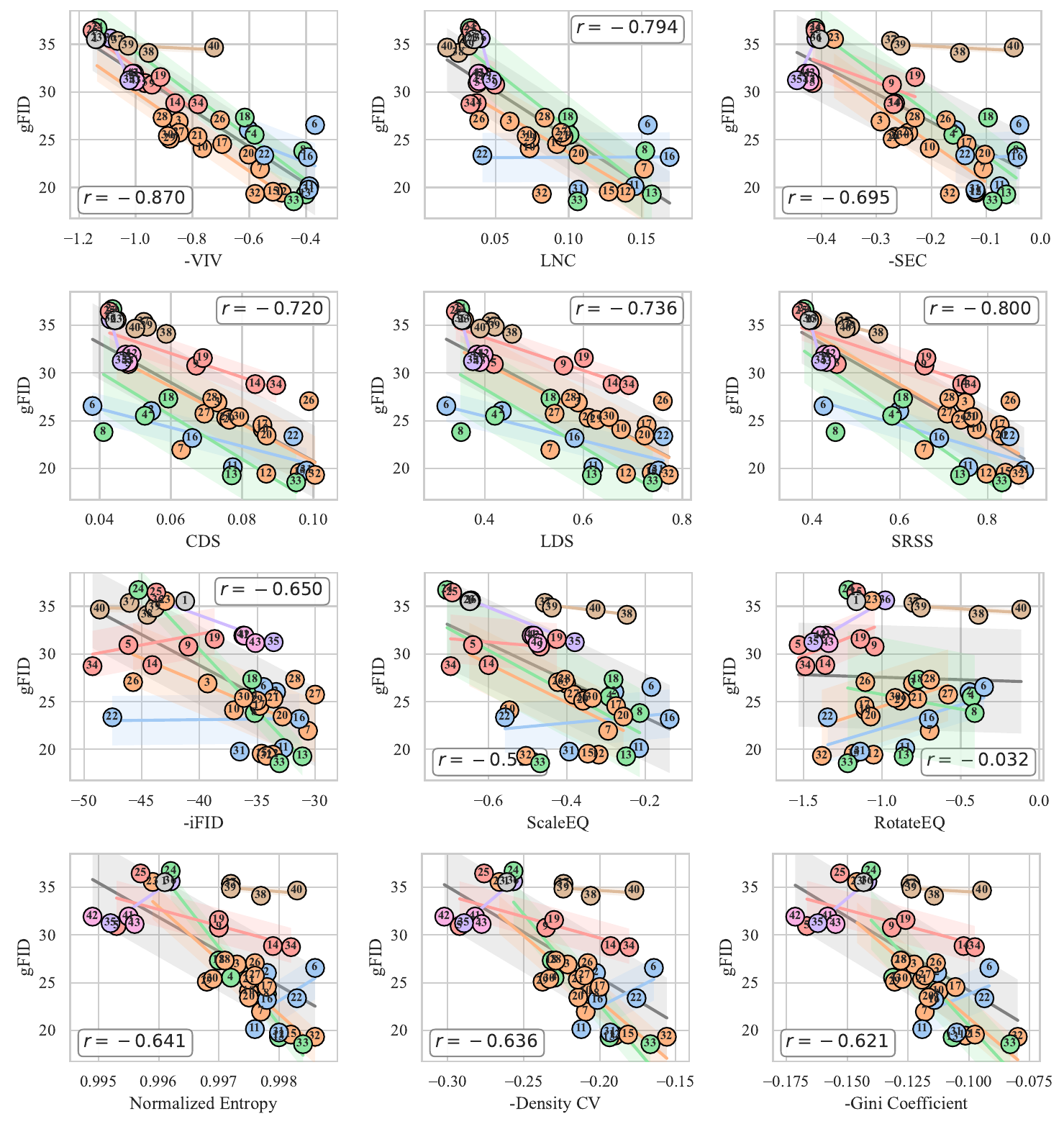}
    \caption{SiT-B gFID with convolutional \texttt{f16d32} tokenizer family.}
    \label{plot:gFID_sit_b}
\end{figure}

\begin{figure}
    \centering
    \includegraphics[width=\linewidth]{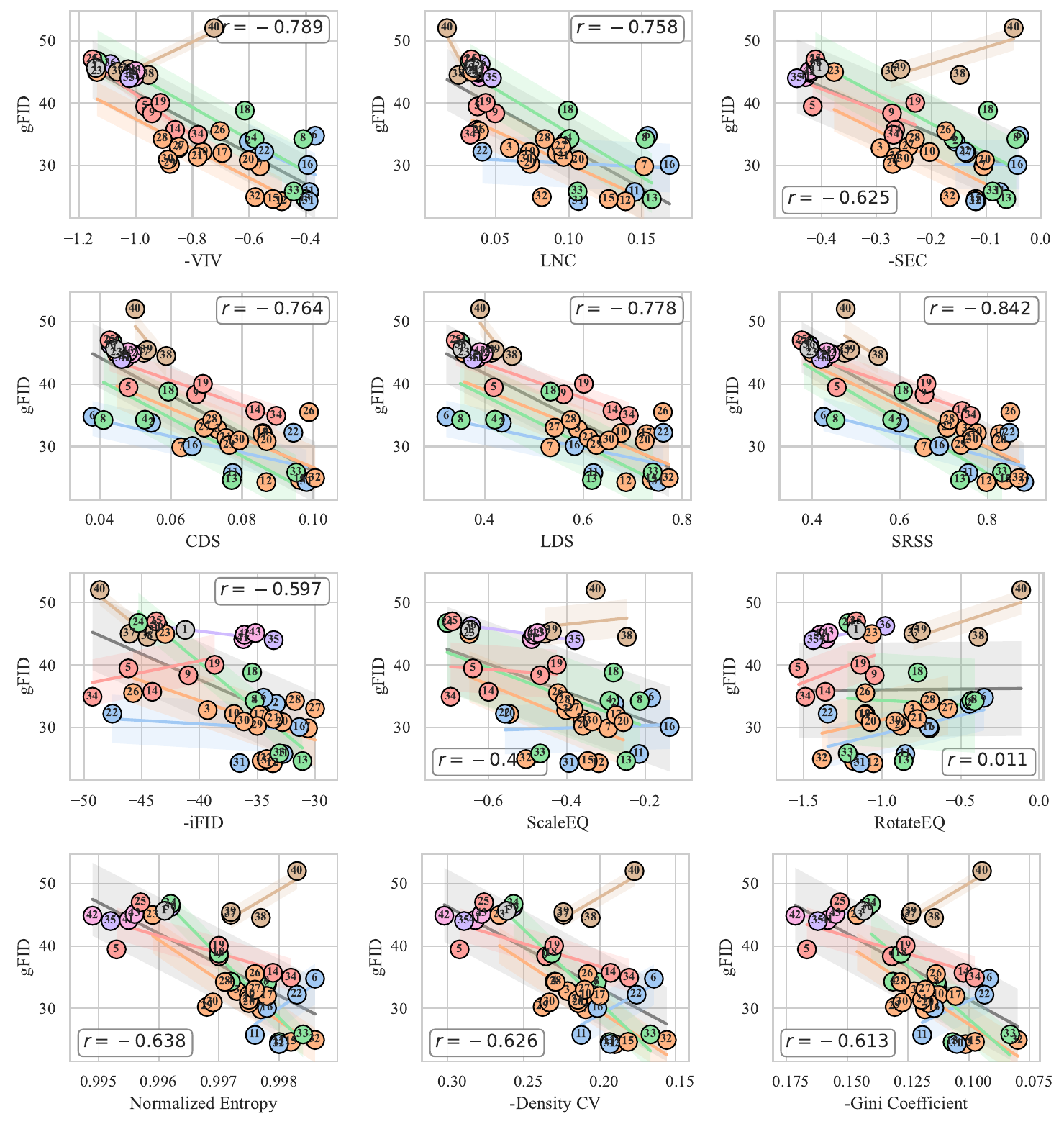}
    \caption{SiT-XL gFID with convolutional \texttt{f16d32} tokenizer family.}
    \label{plot:gFID_sit_xl}
\end{figure}

\begin{figure}
    \centering
    \includegraphics[width=\linewidth]{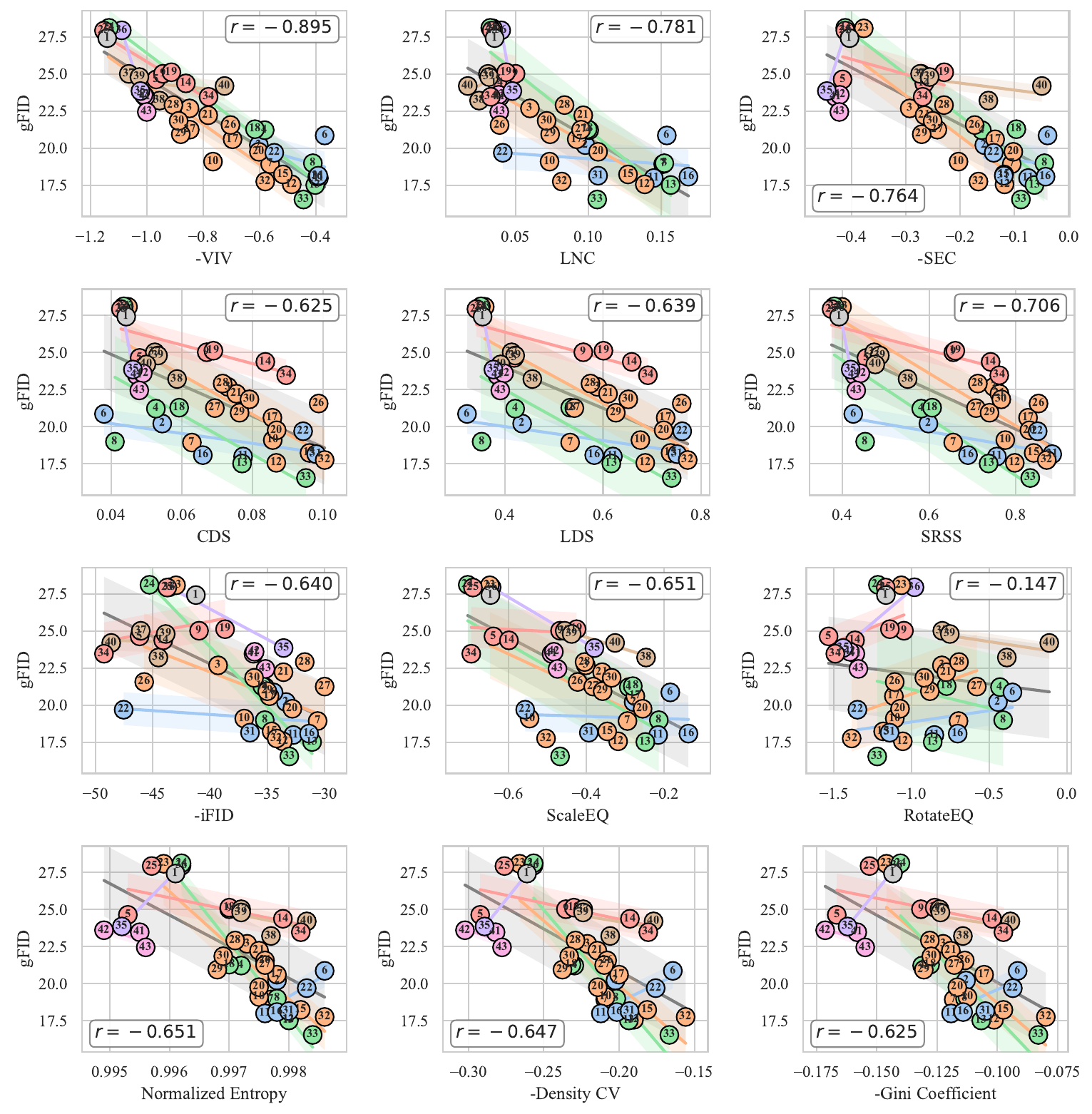}
    \caption{LightningDiT-B gFID with convolutional \texttt{f16d32} tokenizer family.}
    \label{plot:gFID_dit_b}
\end{figure}

\begin{figure}
    \centering
    \includegraphics[width=\linewidth]{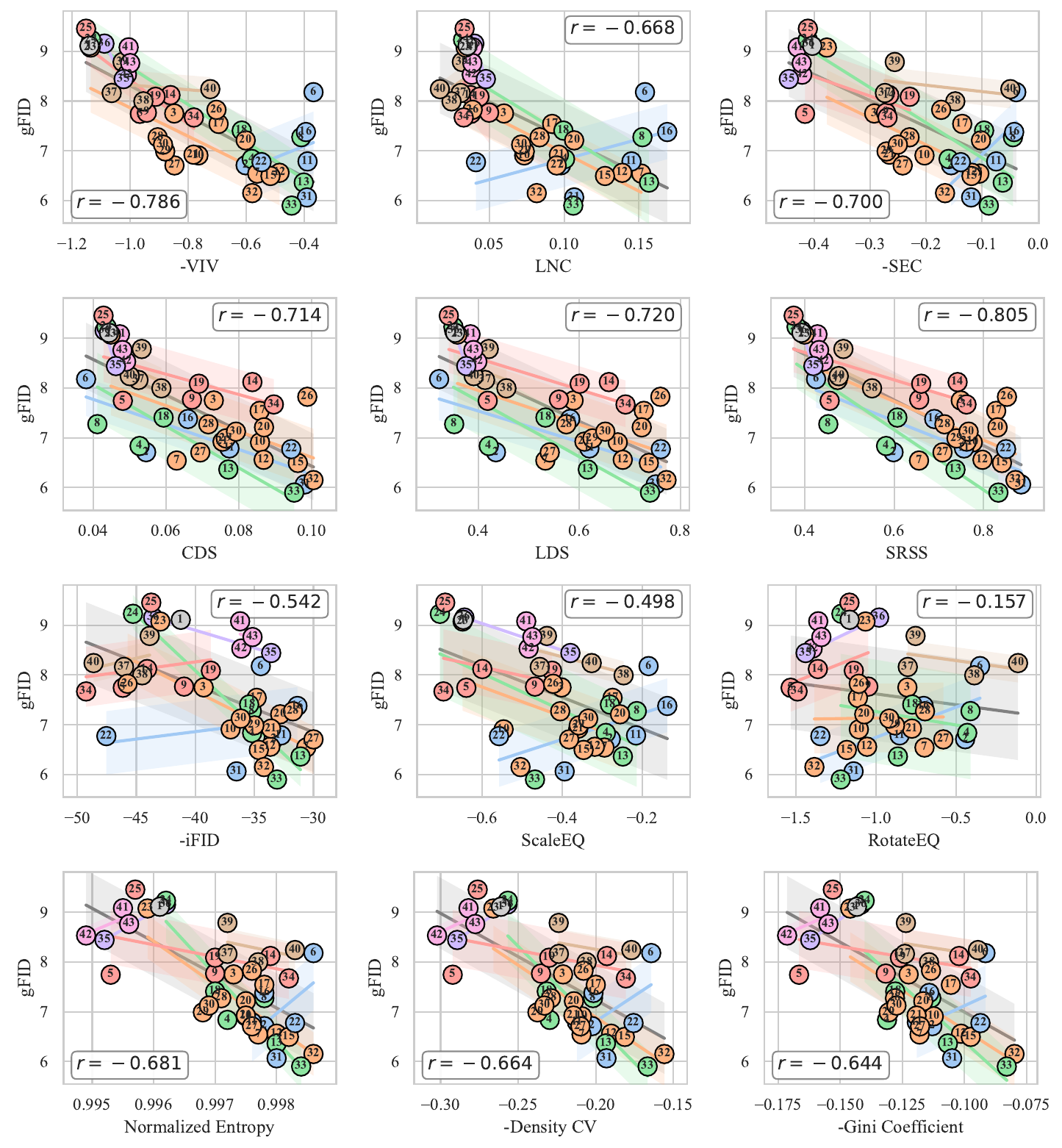}
    \caption{LightningDiT-XL gFID with convolutional \texttt{f16d32} tokenizer family.}
    \label{plot:gFID_dit_xl}
\end{figure}

\begin{figure}[!h]
    \centering
    \includegraphics[width=\linewidth]{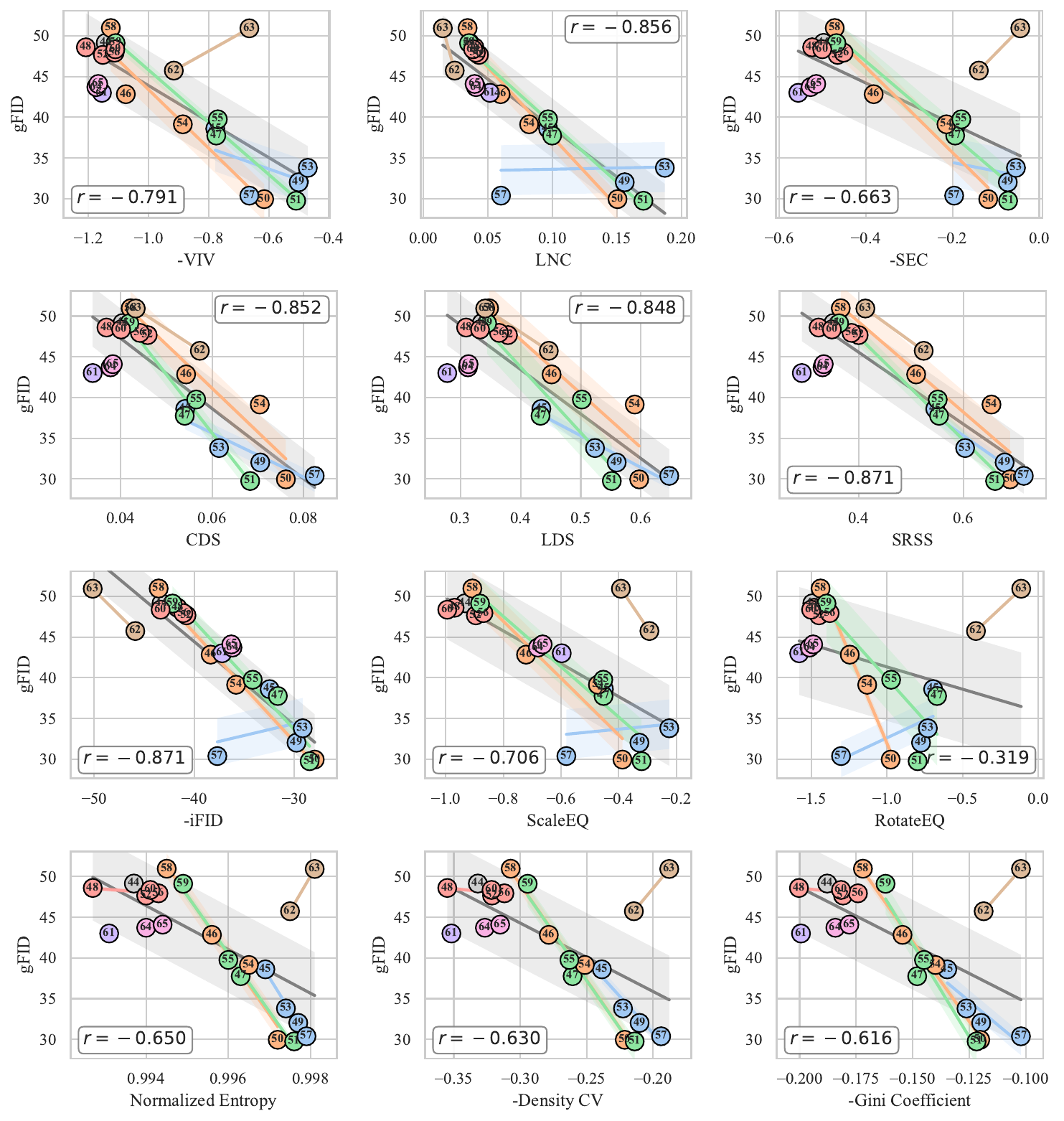}
    \caption{SiT-B gFID with convolutional \texttt{f16d64} tokenizer family.}
    \label{plot:gFID_sit_b_f16d64}
\end{figure}

\begin{figure}[!h]
    \centering
    \includegraphics[width=\linewidth]{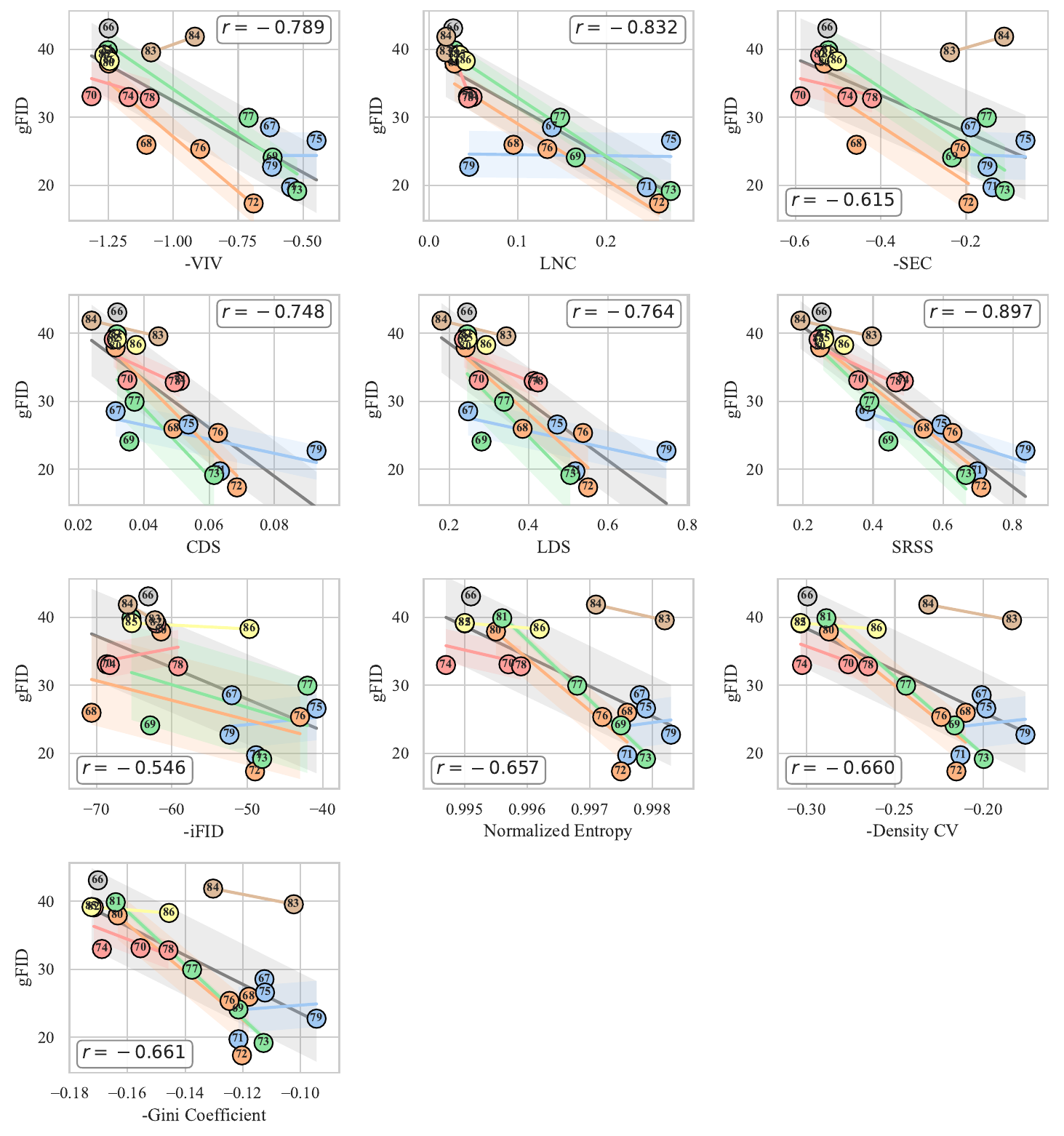}
    \caption{SiT-B gFID with transform-based \texttt{f16d32} tokenizer family.}
    \label{plot:gFID_sit_b_trans}
\end{figure}

\clearpage
\section{Detailed Figures for IS}
\begin{figure}[!h]
    \centering
    \includegraphics[width=\linewidth]{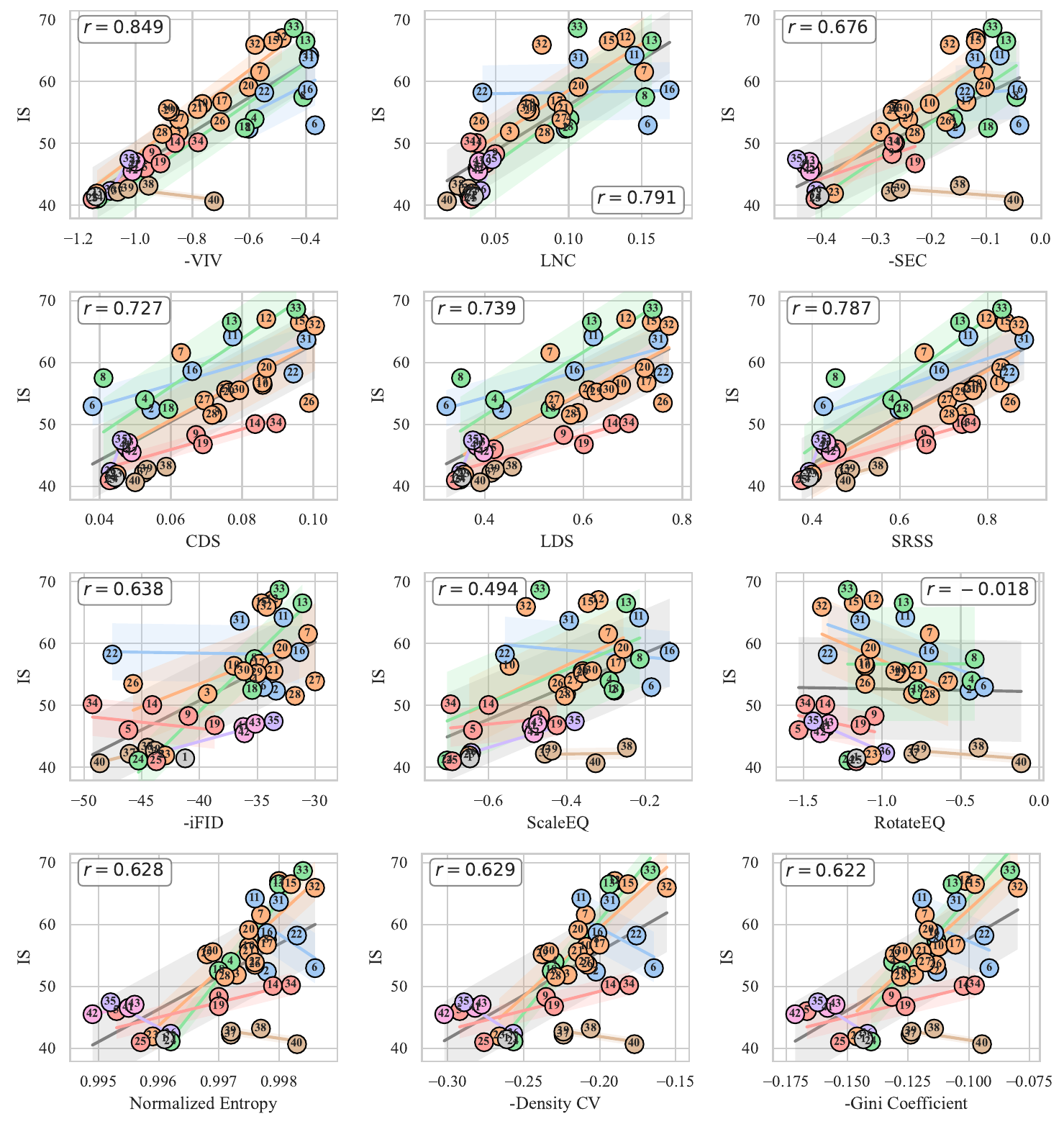}
    \caption{SiT-B IS with convolutional \texttt{f16d32} tokenizer family.}
    \label{plot:IS_sit_b}
\end{figure}

\begin{figure}
    \centering
    \includegraphics[width=\linewidth]{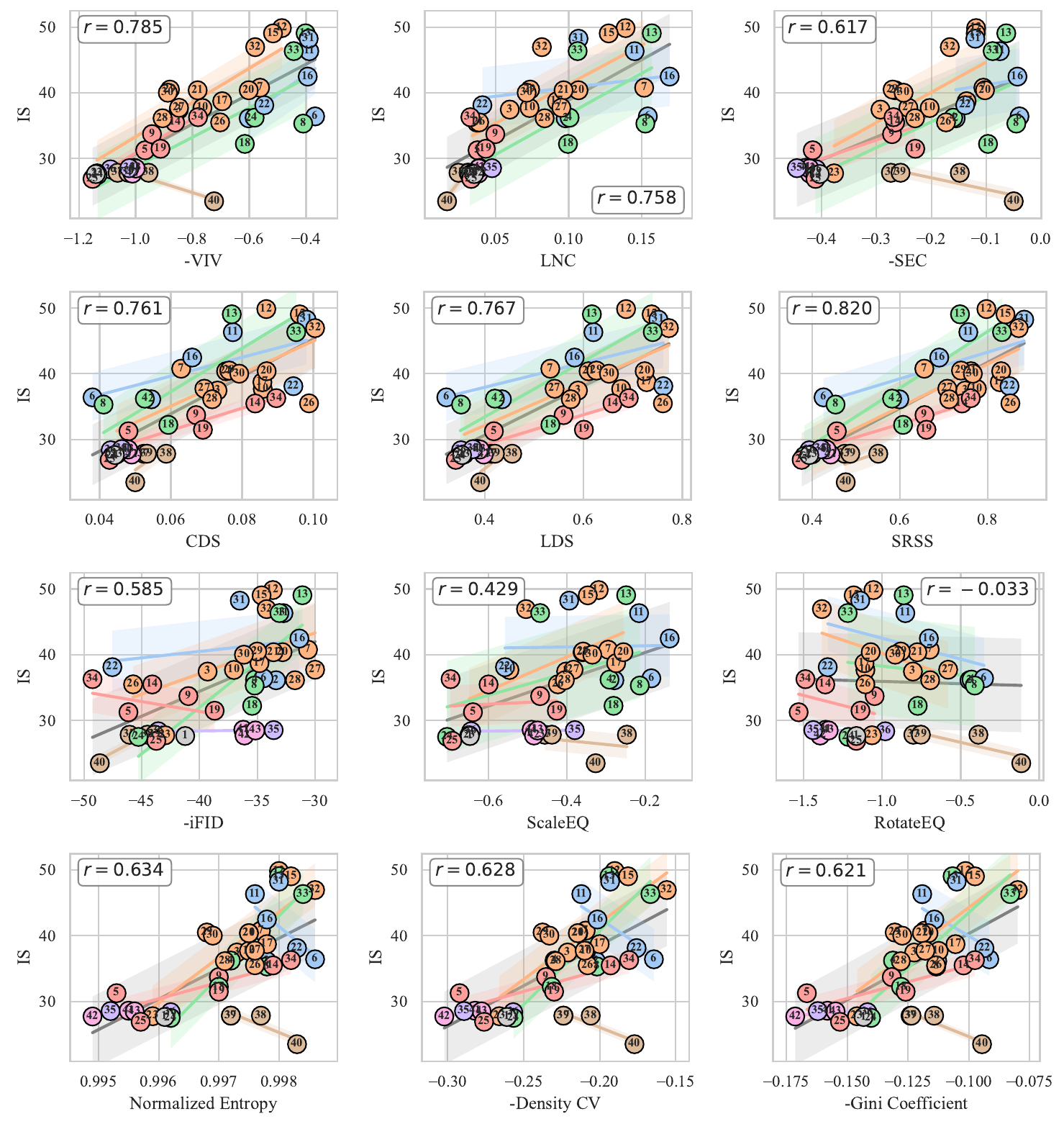}
    \caption{SiT-XL IS with convolutional \texttt{f16d32} tokenizer family.}
    \label{plot:IS_sit_xl}
\end{figure}

\begin{figure}
    \centering
    \includegraphics[width=\linewidth]{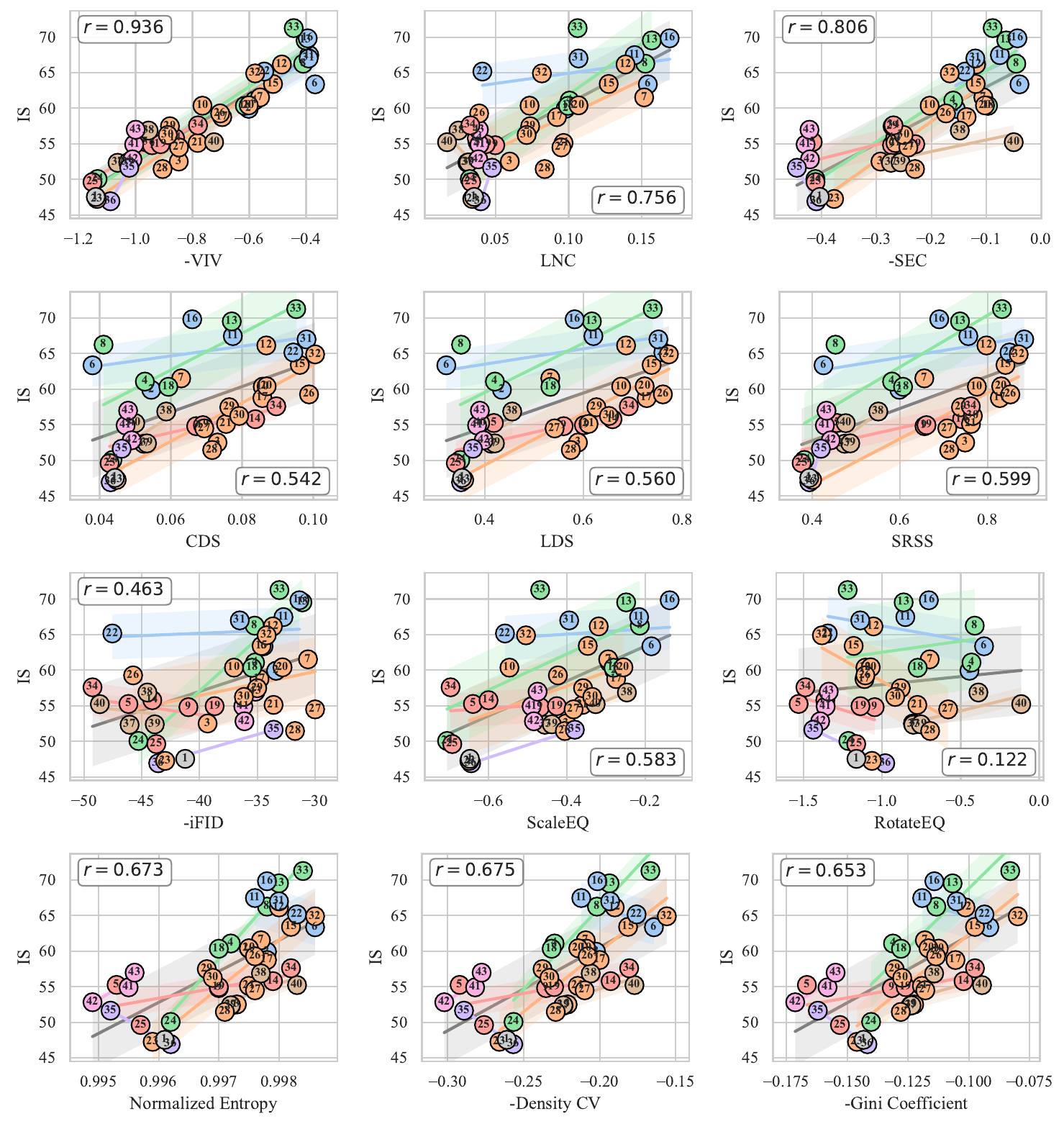}
    \caption{LightningDiT-B IS with convolutional \texttt{f16d32} tokenizer family.}
    \label{plot:IS_dit_b}
\end{figure}

\begin{figure}
    \centering
    \includegraphics[width=\linewidth]{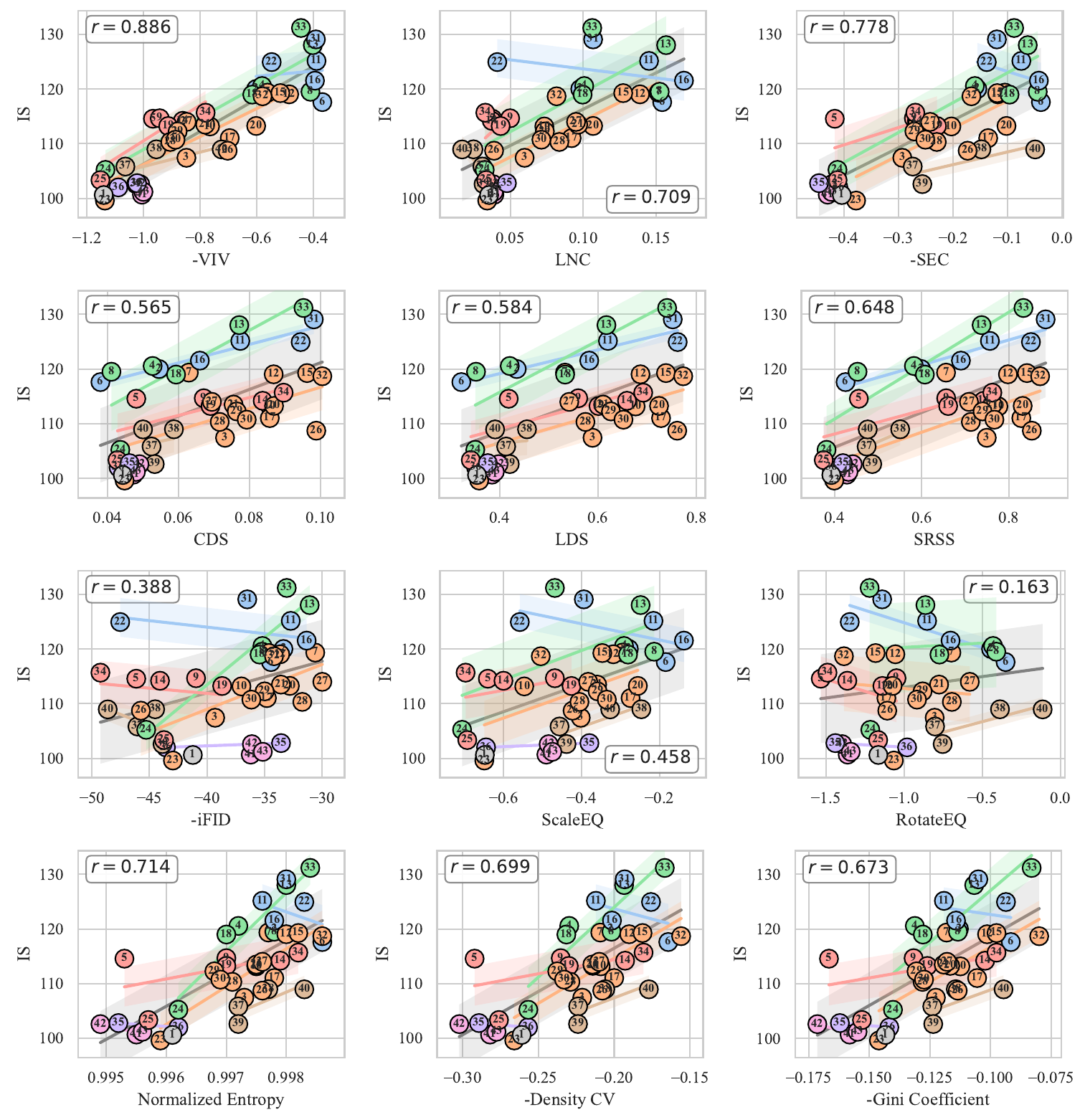}
    \caption{LightningDiT-XL IS with convolutional \texttt{f16d32} tokenizer family.}
    \label{plot:IS_dit_xl}
\end{figure}

\begin{figure}[!h]
    \centering
    \includegraphics[width=\linewidth]{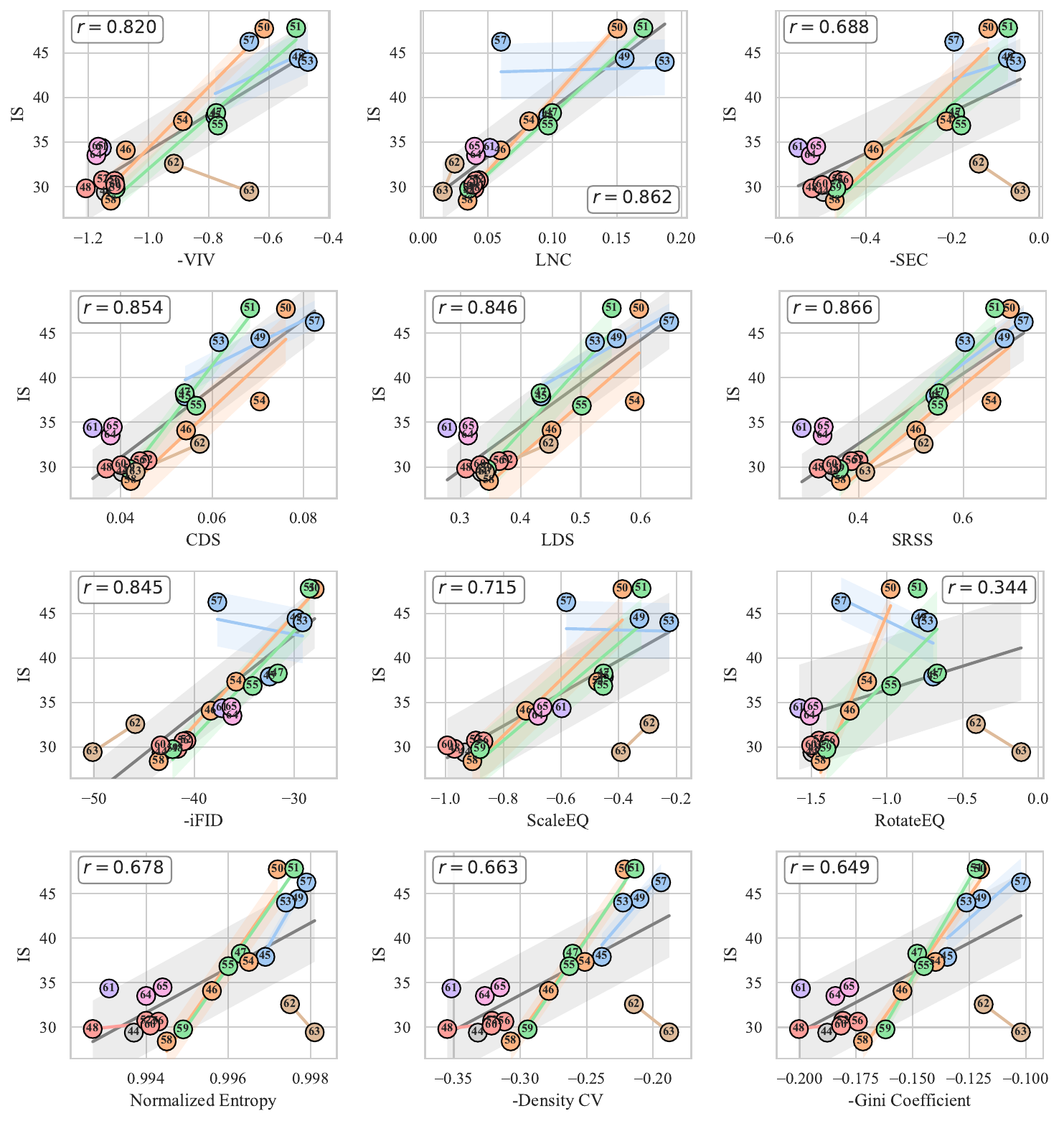}
    \caption{SiT-B IS with convolutional \texttt{f16d64} tokenizer family.}
    \label{plot:IS_sit_b_f16d64}
\end{figure}

\begin{figure}[!h]
    \centering
    \includegraphics[width=\linewidth]{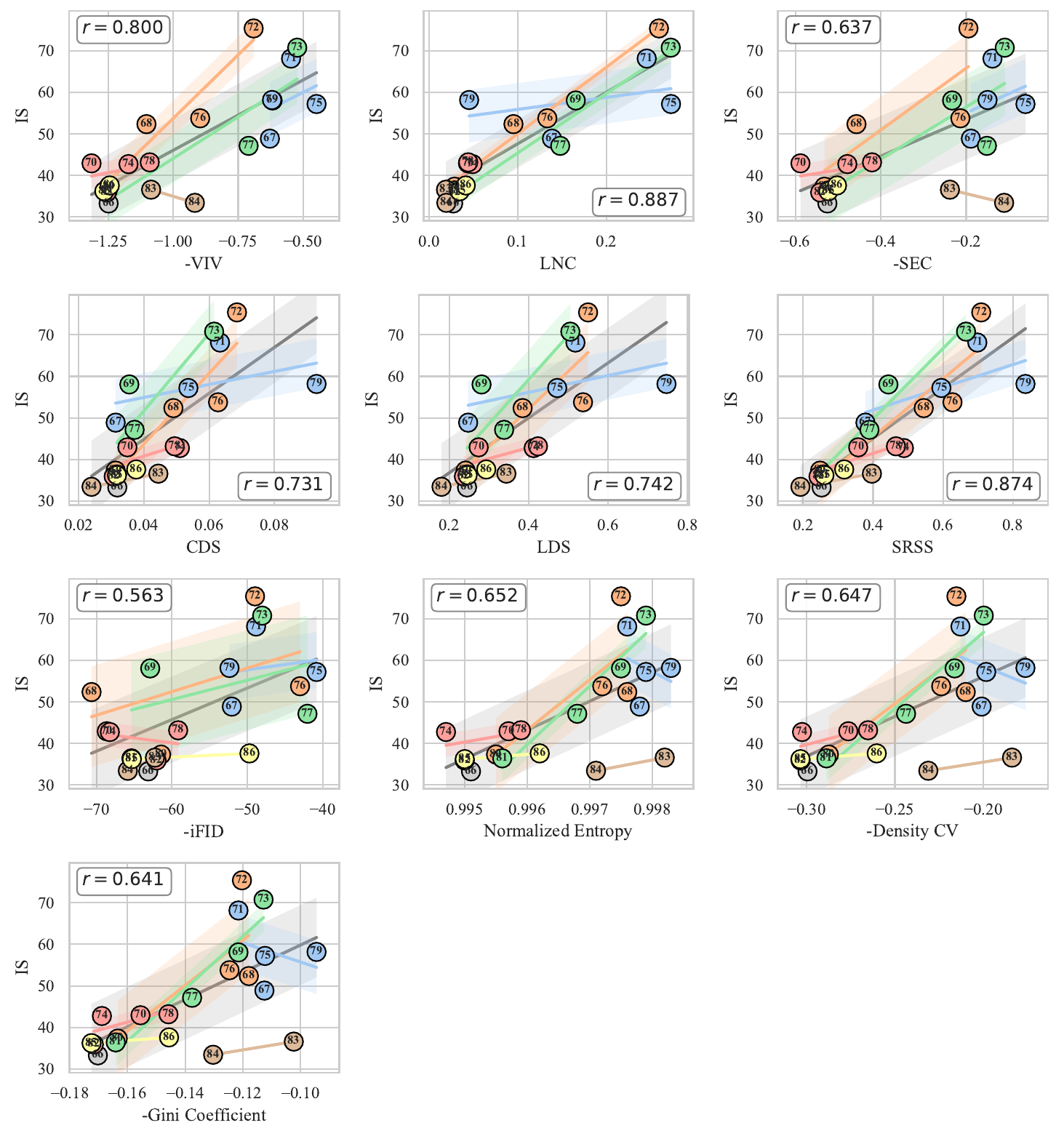}
    \caption{SiT-B IS with transformer-based \texttt{f16d32} tokenizer family.}
    \label{plot:IS_sit_b_trans}
\end{figure}

\clearpage
\section{Detailed Figures for FDr$^6$}
\begin{figure}[!h]
    \centering
    \includegraphics[width=\linewidth]{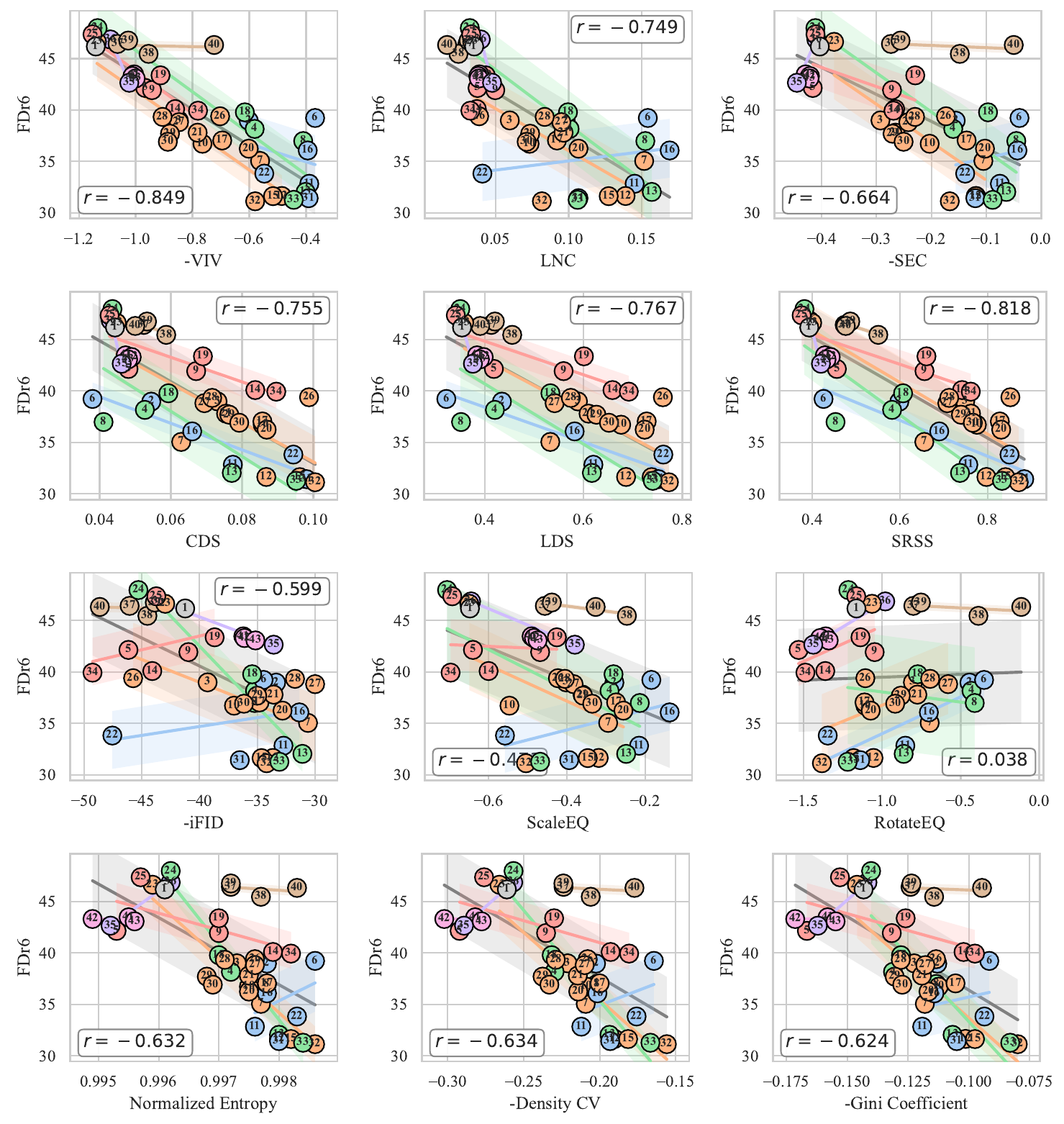}
    \caption{SiT-B FD$^6$ with convolutional \texttt{f16d32} tokenizer family.}
    \label{plot:FDr6_sit_b}
\end{figure}

\begin{figure}
    \centering
    \includegraphics[width=\linewidth]{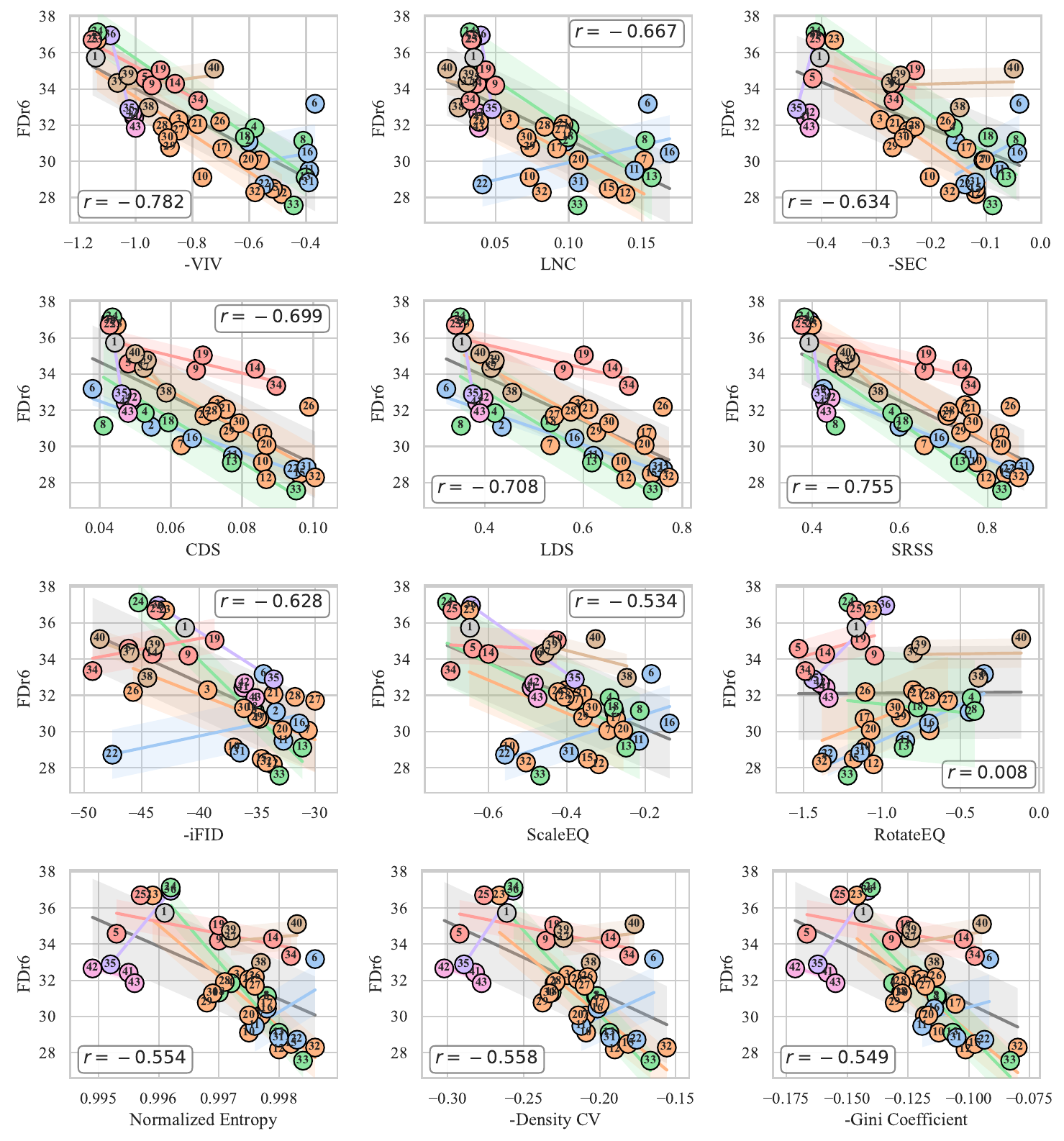}
    \caption{LightningDiT-B FDr$^6$ with convolutional \texttt{f16d32} tokenizer family.}
    \label{plot:FDr6_dit_b}
\end{figure}

\clearpage
\section{gFID Curves with Various CFG Scales}
\begin{figure}[!h]
    \centering
    \includegraphics[width=\linewidth]{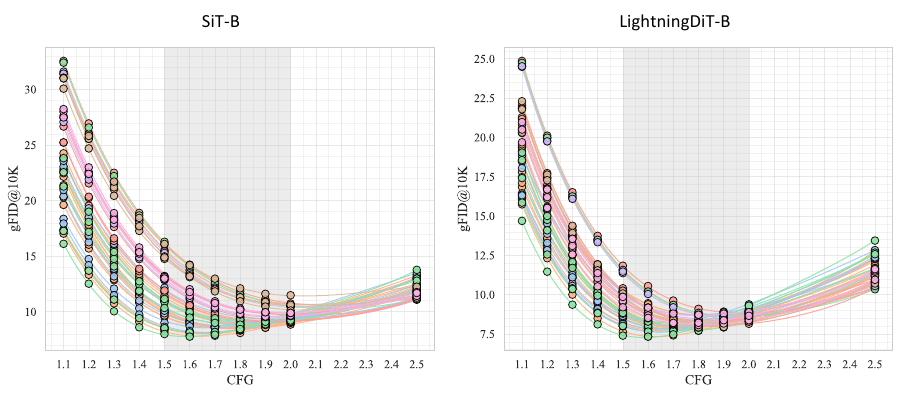}
    \caption{The variation of gFID with CFG for different tokenizers, where the optimal CFG is within the range of 1.5 to 2.0.}
    \label{plot:cfg_curve}
\end{figure}

Figure~\ref{plot:cfg_curve} shows the trend of generation quality on SiT-B~\cite{ma2024sit} and LightningDiT-B~\cite{yao2025vavae} as a function of CFG~\cite{ho2022cfg} scale for different tokenizers.
This shows that the optimal CFG across all approaches lies between 1.5 and 2.0, therefore, Figure~\ref{plot:cfg} presents the results for these sample points.
Meanwhile, we also observed an overall trend that the optimal CFG scales of the better generation approaches are smaller.

